\documentclass[final,3p,times]{elsarticle}

\usepackage{amssymb}
\usepackage{amsmath}

\usepackage{hyperref}
\usepackage[dvipsnames]{xcolor}
\usepackage{tcolorbox}

\usepackage[normalem]{ulem}

\usepackage{setspace}\usepackage{colortbl}%
  \newcommand{\myrowcolour}{\rowcolor[gray]{0.925}}
  \newcommand{\ra}[1]{\renewcommand{\arraystretch}{#1}}
\usepackage{booktabs}

\usepackage{listings}
\lstdefinestyle{yaml}{
     basicstyle=\color{blue}\footnotesize,
     rulecolor=\color{black},
     string=[s]{'}{'},
     stringstyle=\color{blue},
     comment=[l]{:},
     commentstyle=\color{black},
     morecomment=[l]{-}
 }

\usepackage{subfig}
\newcommand{\subf}[2]{%
  {\small\begin{tabular}[t]{@{}c@{}}
  #1\\#2
  \end{tabular}}%
}

\journal{Reliability Engineering \& System Safety}

\begin{document}

\begin{frontmatter}



\title{A Local Agentic AI Framework to Generate Edge Case Scenarios for Safety Assurance}


\author[1]{Victoria J. Hodge\corref{cor1}}
\ead{victoria.hodge@york.ac.uk}
\cortext[cor1]{Corresponding author}

\author[1]{Colin Paterson}
\ead{colin.paterson@york.ac.uk}

\author[1]{Rob Alexander}
\ead{rob.alexander@york.ac.uk}

\affiliation[1]{organization={Department of Computer Science},
            addressline={University of York}, 
            city={York},
            postcode={YO10 5DD}, 
            country={UK}}

\begin{abstract}
Generating edge case scenarios is crucial for training, testing and safety assuring autonomous systems. This identifies hazardous failures to increase system reliability and robustness, ensures the system can still function safely on the boundaries of its operational design domain and learns the expected system behaviour for these edge cases. The rapid advancement in the generation and analysis capabilities of language models (LMs) has expanded their applicability and recent work has investigated LM scenario generation. However, many recent approaches rely on human-in-the-loop or rule-based strategies. This restricts the LM’s ability to generate diverse and edge case scenarios; the long tail of the distribution. Our work uses local LMs in a laptop-based agentic AI framework for adaptive scenario generation. This is cheap, fully-controlled, ensures data privacy and security. It uses think, act, observe, update, and continue loops with lightweight JSON encoding to generate scenarios. We use the system requirements, system description and environment description as knowledge bases in contextual prompts for an LM to ensure scenario realism. A second, independent LM critiques each scenario using domain specific criteria so it can be refined. The refined scenarios are run in simulation and evaluation metrics collected. These metric scores are analysed and similar scenarios are created for edge cases (exploitation) or novel scenarios are generated to explore the scenario space.  Post-hoc clustering using metric scores and thematic-type analysis of the rich-text scenario descriptions produced by the LM pinpoints functional gaps and their triggering conditions. This creates traceable evidence for the safety assurance case. Our work highlights the potential of a flexible and adaptive local agentic AI framework for automating edge case generation and producing traceable safety assurance evidence.
\end{abstract}


\end{frontmatter}


\section{Introduction}
\label{sec:Introduction}
The impact of recent innovations in AI systems has changed how businesses and even economies function and changed how people interact with their environment. The emergence of AI-based autonomous systems is a major part of this innovation. An autonomous system is: ``\textit{a system involving software applications, machines, and people, that is able to take actions with little or no human supervision}''\footnote{https://tas.ac.uk/our-definitions/ (accessed 27 Sept 2026)}. These systems have the potential to revolutionise daily life and their use is expanding rapidly, in particular in safety-critical domains. Safety-critical domains include autonomous road vehicles; medical, assistive, manufacturing and delivery robots; aerial vehicles; marine vehicles; underwater vehicles;  wireless sensor networks used for monitoring; and components within systems such as an autopilot system in an aircraft \cite{hodge2025outofdistributiondetectionsafetyassurance}.  

The development of autonomous systems that operate in safety-critical domains remains complex. Failures in such systems can cause harm to the environment, harm to people, or substantial economic loss~\citep{bozzano2010design}. As systems become more autonomous, the level of human oversight decreases \cite{porter2025INSYTE}. Replacing humans with AI in these systems means that we can no longer rely on a suitably qualified human to respond to unusual situations in a safe manner. Instead, we are required to argue that, throughout its operational lifetime, the system will, autonomously, meet operational, safety, and regulatory standards under all anticipated operating conditions. However, such safety assurance requires rigorous and often expensive through-life safety assurance processes to ensure safety standards and regulations are met, and that the system is free from unacceptable risks. Autonomous systems that operate in open-world environments shift the focus of safe development and operation from guaranteeing against known operational parameters to effectively anticipating and mitigating novel inputs, unexpected situations and uncertainty. Furthermore, the system's operational environment may be underspecified at design time \cite{porter2025INSYTE}, or even unknown, which further increases the likelihood of unknown and unsafe scenarios. Arguing that the system is free from unacceptable risk is a complex and multifaceted process.

\subsection{Problem Statement}
\label{sec:problem}
The required safety guarantees on the autonomous system that must be evidenced to argue its safety can be represented as $g(\mathbf{AS})$ and include requirements such as minimum obstacle clearance, minimum safe lighting level, or task completion time limits. An operational design
domain (ODD) specifies this operational environment within which an autonomous system is intended
to operate reliably and safely. The ODD effectively defines the parameters for system testing \cite{ryan2024will}. Systematic and comprehensive testing throughout development and operation is vital for ensuring the system will operate reliably and safely within its ODD.   Scenario-based testing tests systems against meaningful scenarios and can be used during training and verification and validation of AI models to ensure robustness and provide evidence that $g(\mathbf{AS})$ will be met. There are three levels of scenarios for safety assurance testing in decreasing level of abstraction: functional (linguistic description), logical (range and distribution of parameters), and concrete \cite{iso21448,karunakaran2023generating}. Concrete scenarios $\theta$ are sampled from the logical scenario distribution $\Theta$ and specify a value for each parameter. These concrete scenarios can then be  sub-divided into typical, critical (safety risk present) and edge-cases~\cite{karunakaran2023generating}. Edge cases are statistical outliers \cite{hodge2011outlier} that are rarely encountered in normal operational activities and lie on the boundaries of the ODD. We do not consider these three classes to be mutually exclusive as typical scenarios may be critical and edge cases may be critical. Typical scenarios can be rapidly generated using methods such as random generation where ensuring coverage is the key. Critical scenarios can be targeted by random generation in conjunction with approaches such as fuzzy logic \cite{proma2026fuzzifying} and could fall under typical or edge case. 

Modern safety standards for autonomous systems such as ISO 21448: safety of the intended functionality (SOTIF) \cite{iso21448} and PAS 8800 \cite{iso8800} require specification of spatial edge case scenarios. The wide operating context of autonomous systems means the parameter space of $\Theta$ is vast and the parameters have complex, non-linear relationships. Identifying the parameters $\theta$ to describe realistic edge case is complex and challenging. For example, placing an obstacle in a specific location may only cause a hazard if other obstacles are placed in specific locations. ISO 21448 clause 7 requires us to identify triggering conditions such as specific environmental layouts that expose functional gaps and lead to hazardous events. Clause 8 considers residual risk and introduces principles such as ALARP (as low as reasonably practicable) to reduce risk until any  further reduction would cost far more than the safety benefit. Clause 9 mandates developing a verification strategy so we can find and mitigate issues and clause 10 mandates verification of the known scenario space to argue that known hazards are handled cleanly. We need to consider clause 11 which mandates validation of the unknown scenario space using testing approaches such as simulation and randomized scenario scaling. As the operational design domain of autonomous systems is vast, testing is limited by the resources available so needs to be focused where it is most useful and the ODD specification, whether specified by text or as a model, may not fully capture every scenario possible.  Humans can provide linguistic descriptions and specify parameters ranges and distributions but producing a comprehensive set of concrete edge case scenarios by hand is infeasible due to this vast parameter space. Classical scenario generation for autonomous navigation such as Monte Carlo sampling struggles with high-dimensionality and does not target edge cases. For safety assurance, we are particularly interested in finding critical edge cases so we can mitigate the risks. Finding these safety-critical edge cases and testing them can lead to design, ODD or algorithm refinements and demonstrates that functional gaps have been considered and any found have been mitigated.

\subsection{Proposed Solution}
\label{sec:solution}
\begin{figure}[t]
\centering
\includegraphics[width=\linewidth]{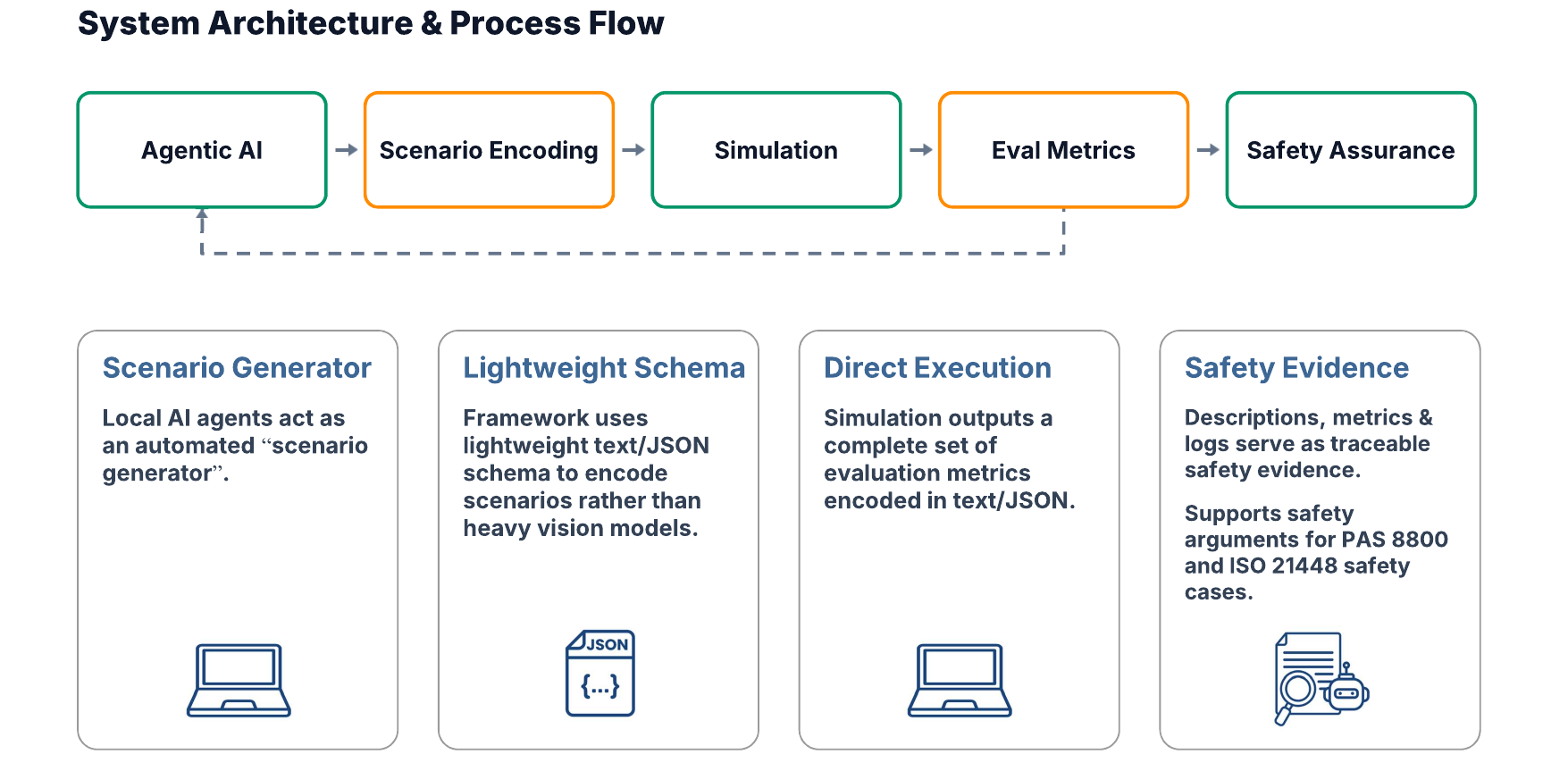}
\caption{The overall architecture and process flow of our proposed agentic AI framework. Icons from \url{https://thenounproject.com/},  the Noun Project}\label{fig:overallArch}
\end{figure}

To address the issues highlighted above we propose the use of a language model (LM) supported scenario generation and testing environment able to identify edge cases for autonomous systems testing in complex environments.
Figure~\ref{fig:overallArch} illustrates the proposed solution which uses agentic AI running on the laptop to programmatically convert safety requirements into structured JSON, hierarchical text-based encodings of scenarios, for stress-testing autonomous systems in simulation. JSON is lightweight and human-readable and is often used for data storage and exchange. The simulation outputs metrics in JSON format to describe how effective each scenario is for uncovering functional gaps in the system's algorithm(s). These scenarios and metric analyses provide traceable evidence for safety assurance.

\subsection{Key Contributions}\label{sec:key-contributions}

\begin{itemize}
\item A programmatic mapping framework translating formal ISO 21448 and PAS 8800 safety specifications into JSON-based spatial coordinate schemas.
\item An iterative and closed loop scenario generation framework capable of dynamic improvement of edge case scenario quality.
\item Empirical validation in simulation demonstrating higher edge-case yield and hazard discovery rates compared to baseline generation techniques.
\item Integration of simulation outputs as verifiable evidence artefacts for safety assurance cases.
\item The agentic AI framework demonstrates that such technology is feasible for smaller companies and ordinary users as it can operate on consumer-grade hardware (laptop).
\end{itemize}

The remainder of this paper reviews relevant safety standards and the scenario-based testing literature in Section \ref{sec:Background}, details and motivates the architecture of our framework in Section \ref{sec:sysArch}, describes how the framework relates to  formal safety requirements in Section \ref{sec:formal}, describes our empirical analyses in Section \ref{sec:experiments},  with the results in Section \ref{sec:results}. We evaluate and discuss the analyses in Section \ref{sec:discussion} along with identified limitations, and Section \ref{sec:ConclusionFuture} draws conclusions and identifies relevant future work to further validate and improve the system.

\section{Regulatory Background \& Related Literature}
\label{sec:Background}
Many previous testing approaches for the safety assurance of autonomous systems are criticised for taking a component-level approach by stress-testing  individual robotic components \cite{pandy2025advancing}. Taking a system-level approach and testing the whole system in operation requires us to identify the optimal set of parameters to accurately describe all aspects of a system, its environment, tasks, and expected behaviour. This requires considerable effort and needs an automated, compute-efficient method to be tractable. Scenario-based testing is a vital part of this process.
Many practitioners take a three step approach to scenario-based testing: generate scenarios, simulate and analyse. Next, we discuss the safety assurance requirements for scenario generation and testing and how scenario-based testing provides evidence for demonstrating $g(\mathbf{AS})$ have been met in the safety case. We consider the simulators used for scenario evaluation and then discuss the relevant scenario generation approaches in the literature.

\subsection{Safety Assurance Standards and Frameworks}
\label{sec:frameworks}

As the level of automation in autonomous systems increases, ensuring they are safe to operate in safety-critical domains requires safety assurance methodologies specifically targeted at this task. Traditional safety standards such as ISO 26262 \cite{ISO26262:2018} for functional safety were designed around deterministic software architectures and explicit boundary conditions pre-specified in the ODD. For AI-enabled systems, these standards exhibit clear 
gaps, in particular failing to account for performance gaps rooted in the high-dimensional input spaces of AI-based systems. The operating environment may be underspecified at design time, or even unknown, and the generalisation of design-time verification to run-time safety is problematic which all increase the likelihood of functional gaps \cite{burton2020MindTheGaps}.  Thus, autonomous systems lack well-understood safety guarantees and this can lead to unpredictability and ultimately hazardous failures~\citep{hsu2024safety}. Furthermore, current AI validation approaches rely on empirical performance metrics (such as recall and precision) evaluated over static datasets. This hides local performance failures and fails to quantify the AI model’s capacity for safe interpolation and extrapolation when encountering edge cases or out-of-distribution scenarios \cite{hodge2025outofdistributiondetectionsafetyassurance} in the ODD.

The newer ISO 21448 (SOTIF) \cite{iso21448} lifecycle prescribes procedures for identifying and controlling system risks so the residual risk shall be demonstrated to be below an acceptable level by gathering evidence from safety verification and validation. Further, SOTIF defines four areas based on two metrics; \textit{knowledge of the scenario} and \textit{safety}. It classifies all scenarios along using these two metrics as axes, knowledge (known/unknown) and hazard (safe/unsafe) where unsafe are the critical scenarios defined earlier. System verification focuses on area 2 (known/unsafe: boundary tests to pinpoint where a system transitions from failing to safe) and validation focuses on area 3 (unknown/unsafe: edge cases that are real-world situations that system engineers have not tested). Thus, identifying boundaries between safe and unsafe and identifying edge case scenarios to reduce their occurrence within the ODD is an urgent problem in safety verification and validation \cite{wang2024} and is the focus of this research. Edge cases have a \textit{Triggering Condition} (SOTIF Clause 7) which represents a specific set of parameters (physical properties and environmental states) that cause a functional gap where system performance does not meet desired performance. They can be identified by system-level safety analysis, component-level failure analysis, fault trees, and simulation. Scenario generation tools coupled with simulation are designed specifically to combine parameters until a (functional gap) triggering condition occurs. \cite{hodge2025agile} note the importance of scenario-based testing during the development lifecycle for assuring the safety of autonomous systems.

PAS 8800 (Road Vehicles - Safety and Artificial Intelligence) \cite{iso8800} complements ISO 21448 with a normative safety framework for managing the AI safety lifecycle. While PAS
8800 defines the necessary safety objectives and process requirements, it remains strictly
non-prescriptive regarding implementation mechanisms. PAS 8800 specifies what must be achieved
to ensure safety but leaves an open methodological gap regarding how these requirements
can be verified, quantified, and traceably validated in practice. While SOTIF testing identifies triggering conditions, PAS 8800 mandates that synthetic edge-case scenarios must be generated to identify data gaps, AI model gaps, non-deterministic behaviour and to validate the AI model's generalisation and robustness boundaries \cite{iso8800}.

STAMP (System-Theoretic Accident Model and Processes) is a safety framework based on systems theory and STPA (System-Theoretic Process Analysis) is a hazard analysis method derived from STAMP \cite{leveson2012engineering}. Together they can generate scenarios \cite{tsuji2020}. However, the number of scenarios generated can be large which is ideal for ensuring scenario coverage but not for targetting edge cases as only a small portion of the scenarios generated will be edge cases (the long tail of the distribution). \cite{tsuji2020} use statistical model-checking to search the STAMP/STPA scenario space for critical scenarios (those likely to cause hazards). However, the approach requires specification of UML and statistical modelling templates which are bespoke to the application and not generalisable. Our approach would complement STAMP/STPA by supplementing it with targeted edge cases.

SACE (Safety Assurance of Autonomous Systems in Complex Environments) \cite{hawkins2026guidancesafetyassuranceautonomous} is an eight-stage methodology to safety assure autonomous systems considering their operational domain. Within SACE, scenario generation is important for identifying hazardous interactions  with the operational environment in stage 2 to discover risks, for boundary testing using edge-case scenarios in stage 7 and for verifying system behaviour in stage 8 using simulation and testing pipelines to generate the necessary evidence for a justified safety case. SACE encourages the use of simulation-based testing using a scenario coverage approach but provides little guidance as to how to do it.

\subsection{Robotic Simulation \& Virtual Testing}
\label{sec:simulation}

Simulation-based testing is vital to overcome the testing bottleneck as comprehensively testing autonomous systems across their operational domain is highly complex and computationally expensive. Simulation provides a secure and fully-controlled environment to test the scenarios generated including testing edge cases without risking human life, the environment or property. \citep{kargar2024emerging,choi2021} survey and critique simulation  of autonomous robots focusing on the technologies used and how simulation can ensure the safety, reliability and robustness of the robots.

We can test in simulation and then transition to real-world testing. This requires the simulation to accurately represent the real-world to ensure the simulation testing is valid. Simulations operate at different levels of fidelity. However, \cite{sotiropoulos2017bugs} show that even low fidelity simulation can unearth bugs in a robot's navigation system.  Popular robotic simulation environments such as Gazebo have been used for scenario analyses \cite{proma2025scaloft}. For higher fidelity, game engines such as Unity-3D can realistically simulate complex and dynamic environments, making them ideal for robotics and vehicle simulation. For example, \cite{karunakaran2023generating} use the CARLA simulator which is based on the Unreal game engine and is aimed at road vehicle simulation and \cite{platt2022comparative} show that Unity-3D game engine scales better, achieves better visuals and is more capable of real-time LiDAR simulation than a traditional robotic simulation environment such as Gazebo.

\subsection{AI-Driven Scenario Generation}\label{sec:scenario-generation}

Scenario generation overcomes the data scarcity bottleneck for training AI components of autonomous systems by generating large numbers of scenarios from some more general specification of the scenario space (e.g. a description of the ODD). It can be targeted specifically at edge case scenarios to  challenge the system's safety boundaries and learn optimal behaviour through systematic testing. There is a range of techniques for scenario generation covering typical, critical and edge case scenarios.

\cite{de2017assessment} use Monte Carlo simulations to generate test cases from a parameterised representation. However, the Monte Carlo simulation is a random search and not targetting edge cases. \cite{proma2026fuzzifying} adopt a fuzzy-logic approach which combines human-like descriptions with degrees of membership to target critical scenarios. By using ISO 34504:2024’s tag-based scenario categorisation  \cite{ISO34503} the authors transform an abstract ODD into a concrete, parametrised model describing the possible operating conditions. This motivates an approach that can systematically organize environmental factors, the system description and requirements from the ODD and the task description to represent the combinatorial space of possible autonomous system scenarios. 

Formal methods can use formal analysis and falsification tools to generate test scenario(s) and their safety properties described in a formal language such as SCENIC and the SCENIC scenarios can be executed in simulation and analysed \cite{fremont2020}. \cite{ortega2024} generate a statistical model of the system for testing scenarios and skew the model towards critical scenario generation. However, this modelling requires the robot's behaviour to be accurately statistically modelled across its operational domain which is complex and difficult to capture. 

Authors have used evolutionary and genetic algorithms for scenario generation \cite{althoff2018automatic,zhou2022genetic}. \cite{khatiri2025Surrealist} used a search-based evolutionary algorithm to automatically generate challenging obstacle avoidance scenarios, uncovering failures often missed by manual or random testing. The authors demonstrate that their scenario generation workflow can be integrated into a complex industrial robotics workflow. Their scenarios are similar to ours, laying out obstacles, waypoints and goals. Their workflow generates test scenarios, executes them in simulation, and analyses the results for potential weaknesses. These approaches use random mutations, need extensive configuration and use complex trial and error analyses.

\cite{pandy2025advancing} note that many of the current scenario generation approaches do not provide dynamic adjustment within the scenario generation lifecycle.  \cite{pandy2025advancing,karunakaran2023generating} use dynamic adjustment of test scenarios through reinforcement learning based on observed behaviours, enabling the identification of edge cases and high-risk interactions that are often overlooked in static methods. However, RL does not have the world knowledge of LMs, and requires millions of epochs of training and complex reward specification.

There is a lack of automated, compute-efficient tools that map from high-level safety standards onto executable 3D scenarios specified using standardised formats. There has been a body of work generating scenarios for safety assurance using LLMs but many focus on generating visual scenes \cite{taniguchi2026,xia2026sagescalableagentic3d} and adversarial behaviour simulation \cite{liu2026aaai,Mei_2025}. \cite{Zhao08052026} introduce a pipeline to generate safety-critical scenarios from crash reports and road network diagrams but this requires such datasets to be available. \cite{xia2025failuresfixesllmdrivenscenario} introduce SERA an LLM pipeline to create targeted scenarios from failure cases by analysing performance logs to pinpoint failure patterns and then retrieving semantically aligned scenarios from a pre-existing database of scenarios. \cite{danso2025llms} developed a  semi-automated pipeline that transforms operational design domain (ODD) descriptions (text) into executable CARLA simulation scenarios. They decompose the ODD description using  structured prompt engineering and reasoning into structured scenario specifications which can be run on simulation platforms, such as CARLA.  Their second stage uses a second LLM to refine the scenario using known simulation error patterns and corresponding corrections. They also note that this may require minor manual interventions and adjustments or compensation for missing assets in the simulation. PlannerForge \cite{gao2026plannerforgellmagentsscenariobased} is also a  semi-automated pipeline where the user describes their required scenario in an LLM prompt, the LLM converts this to JSON to systematically describe the user's intent  and creates a scenario using maps and simulated layouts which the user edits. Targeting critical scenarios would require the user to specify critical scenarios in advance.

Our approach is fully automated and does not require prior knowledge of errors, functional gaps or critical scenario examples. Its purpose is to uncover such errors and gaps.
In the following sections, we develop an agentic framework that focuses on edge cases, directly links the ODD to the scenario-generation, does not require prior knowledge of functional gaps or for the entire system and environment to be fully modelled, does not require extensive trial and error testing to fine-tune the scenario generation process, is fully automated and incorporates feedback and dynamic adjustment into the scenario generation process. It is directed by safety assurance processes as we decompose the ODD to a domain ontology with three sections: \{\textit{environment description, system description, system requirements}\} and we also supply a \{\textit{task description}\} for scenario generation. We represent these as knowledge bases (see Section \ref{sec:constraintVal}) which provide information and context in the prompt.

\section{System Architecture \& Local Agentic Closed Loop}\label{sec:sysArch}
\begin{figure}[t]
\centering
\includegraphics[width=\linewidth]{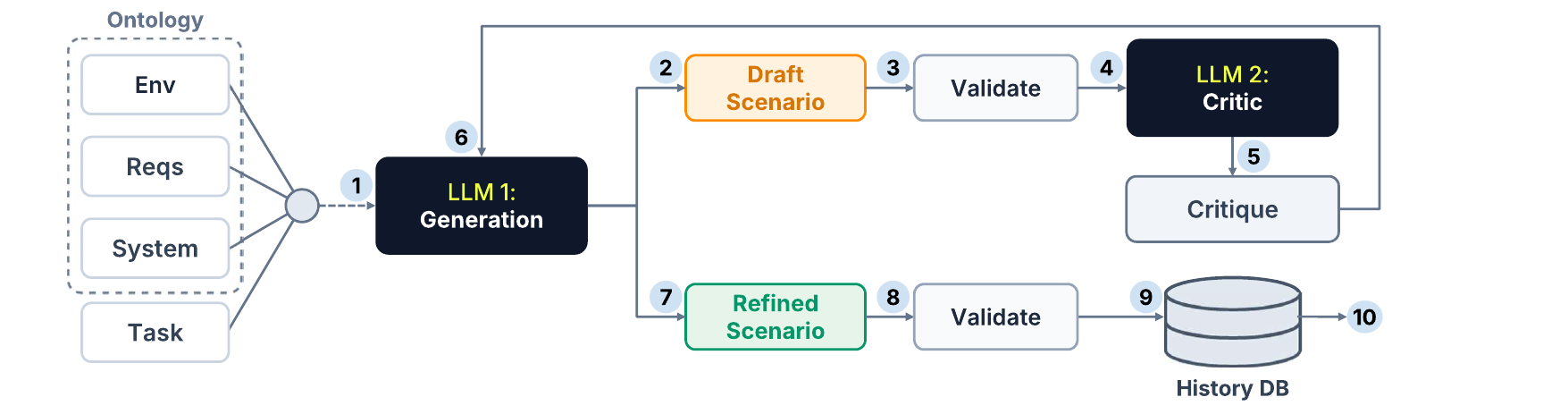}
\caption{Overview of the first part of the agentic AI scenario generation framework. Part 1 produces the JSON-encoded scenarios to run in simulation in part 2 which is shown in Figure \ref{fig:part2}. The domain ontology allows us to map the sections of the operational deign domain description to sections of the LM prompt and is described in section \ref{sec:constraintVal}.}\label{fig:part1}
\end{figure}
\begin{figure}[t]
\centering
\includegraphics[height=8cm]{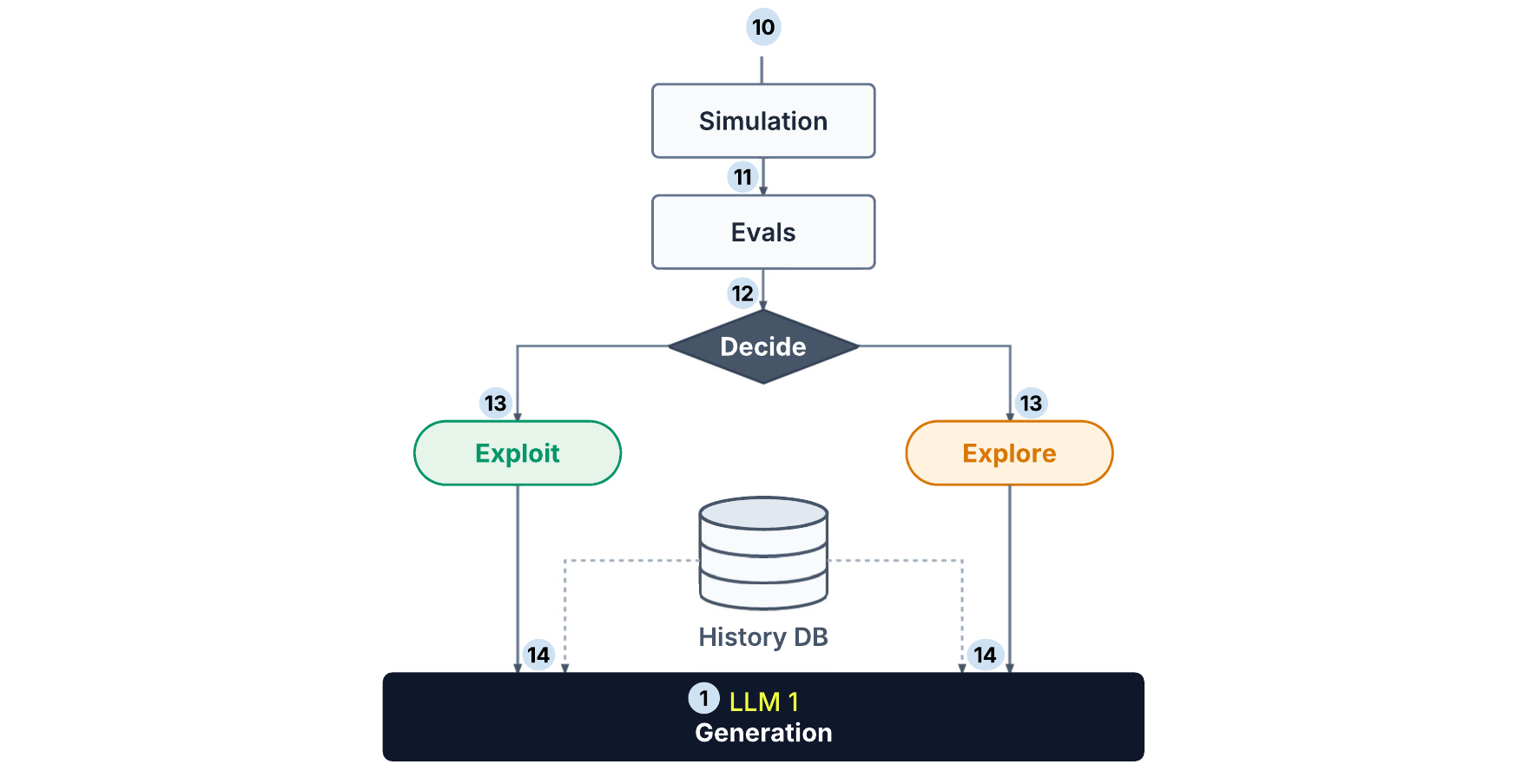}
\caption{Overview of the agentic AI simulation, evaluation and decision logic to assess the scenarios and to decide whether to exploit an existing scenario or explore novel scenarios by looping back to step 1 in Figure \ref{fig:part1}.}\label{fig:part2}
\end{figure}

The agentic loop's task is to search the scenario parameter space for SOTIF triggering conditions derived from formal safety specifications. Triggering conditions are specific  events or spatial layouts that uncover a system weakness that leads to a hazard such as not maintaining safe separation from an obstacle. To find triggers, the lifecycle comprises a number of steps that loop. We illustrate the lifecycle in Figures \ref{fig:part1} and \ref{fig:part2}. In Figure \ref{fig:part1}, step 1) integrates the four sections of our ODD as inputs to an LM, LLM1, which generates 2) draft scenarios and 3) validates them for syntax correctness. A second (different) reflection LM, LLM2, critiques each scenario (4) and LLM1 uses the 5) critique to 6) revise and refine each scenario (7) which is 8) syntactically validated and 9) stored in a scenario history database. In Figure \ref{fig:part2}, step 10 runs each scenario in turn through a simulator and 11) outputs a set of quality metrics (evals), step 12 analyses the metrics and decides whether to 13) exploit or explore, step 14 loops back to the original LM and either exploits (prompts to create more similar scenarios using the current scenario) or explores using the history database in the prompt to generate scenarios that are different from those in the history. 
Once we have finished looping, we have a set of JSON-encoded scenarios and associated evaluation metric results through simulation that form our traceable safety assurance evidence towards showing that $g(\mathbf{AS})$ are met (as described in section \ref{sec:sotifTriggerMapping}). 

We describe how to post-process the generated JSON encoding the scenario and simulation evaluation metrics in sections \ref{sec:qualAnalysis} and \ref{sec:qualSafetyCase}. This uses analyses inspired by thematic analyses \cite{braun2006using} to identify SOTIF triggering conditions, cross-reference the evaluation metric scores for those SOTIF triggers and uses the results as safety assurance evidence. The metrics and corresponding spatial layouts act as concrete, reproducible evidence artifacts. This develops an automated mechanism for transitioning area 3 (unknown/unsafe) scenarios to area 2 (known/unsafe) under SOTIF Clause 7 and provides a platform for transitioning them to area 1 (known/safe) by developing mitigations. 

\subsection{Agentic AI Closed Loop Overview}

Our agentic AI framework is a closed loop that feeds back the evaluation outcomes generated by running the scenarios through the simulator and uses them to improve the next set of generated scenarios thus moving beyond random sampling. It incorporates a ``\textbf{domain ontology}'' which represents the ODD and is specified as a machine-readable model comprising \{\textit{environment description, system description, system requirements,}\}  coupled with a \textit{task description} (see Figure \ref{fig:part1} and section \ref{sec:constraintVal}).  The think, act, observe update and continue loops as described in Section \ref{sec:architectOptimisation} exploit chain-of-thought prompting \cite{wei2023chainofthoughtpromptingelicitsreasoning} by decomposing prompts into step-by-step chunks which reduces logic and reasoning errors in the LMs. We couple this with representing the domain ontology as knowledge bases (see section \ref{sec:constraintVal}) to include structured knowledge into the LM's prompt creating a contextual prompt \cite{bae2026wargames} with background information, specifications, rules, requirements and goals. The  prompts are also schema-strict and force the LM to convert qualitative descriptions (e.g., ``focus on edge cases with dense obstacle clusters'') into quantitative coordinates and environmental layouts in a strict predefined data layout. For our current use case, the loop outputs JSON specifying spatial object placements such as obstacle layouts targeted at SOTIF triggering conditions. Our framework is adversarial as our task description prompts the LMs (see Listing \ref{listing:taskDescription}) to generate long tail scenarios such as complex obstacle patterns or obstacle walls that specifically target likely functional gaps. The generator LM generates the scenario which the critic LM validates and critiques so the generator LM can refine the scenario, the autonomous system attempts this refined scenario in simulation, the simulation provides evaluation metrics to feedback and the generator LM  can use these to refine existing scenarios (exploit) or generate novel scenarios (explore).
  
We use a set of software libraries to develop this agentic framework. AISuite \cite{ng2024aisuite} acts as our overall model client controller as AISuite provides a single, unified interface to different LMs including to different LM providers. This allows our framework to switch between different LMs by simply changing a text string in a line of Python code. This means the framework can be easily modified to work with the user's preferred LMs and can also easily switch between local and cloud-based model providers.

To allow our framework to run as local agentic AI  we use Ollama \cite{ollama2026} which, once installed, provides a simple interface to a variety of LMs. This provides complete data privacy as a users' data never leaves their device, there are no cloud provider usage fees, no per-word cloud costs, and the full framework runs without needing an internet connection. This is important for safety critical environments where safety data may need to remain private.

In the setup in this paper, we use two Ollama LMs. The 
generator LM is \textbf{ollama:qwen3-coder:30b} \cite{cao2026qwen3codernext} and the 
critic/reflection LM is \textbf{ollama:nemotron3:33b} \cite{nvidia2025nemotron3nano}. We chose these models through simple experiments for their accuracy, stability and ability to run on the resources available. These smaller LMs are more task-specific than LLMs so selecting the best models requires careful consideration of each model's abilities (for example ollama:nemotron3:33b is recommended for its reasoning ability so is best suited to reflection) coupled with trial and error testing.

\subsection{Local Agentic Architecture and Optimisation}\label{sec:architectOptimisation}
The local agentic framework described in this paper ran on a Linux Ubuntu 22.04 laptop with Intel i9 CPU, 32GB RAM and a GeForce RTX 4080 GPU.

We considered the design of our agentic framework to optimise its demands on the compute resources so it ran on a single laptop. We treated memory, orchestration, validation, evals, and observability as first-class parts of the system by considering them throughout the development lifecycle. We only use LMs for tasks where they are most useful and use software engineering to implement logic functionality, for example, to decide whether to exploit the current scenario or to explore novel scenarios.

Using a structured text representation for scenarios, such as JSON, minimises the communication overhead between LMs and between LMs and simulation compared to using images or models. For a local agentic framework, this is vital to minimise the demands on the compute resources available and allows closed-loop execution to run alongside Unity-3D simulation on a single GPU laptop. Generating multiple scenarios from one prompt (Step 1 in Figure \ref{fig:part1}) ensures a variety of scenarios will be produced. However, testing indicated that attempting to generate too many from one prompt leads to the Qwen LM used here generating errors in the JSON as working memory is  smaller in local models and they drift and forget to match brackets, for example. 
We generate JSON-encoded scenarios in batches of four, i.e., request four scenarios from the LM as this is the highest value that leads to error-free scenarios. We also use the Python package Pydantic and our Python software to validate all generated JSON and all data types (such as ensuring coordinates are integers) and to fix simple JSON syntax issues such as missing closing braces before the JSON-encoded scenario is executed in simulation. Running the simulation is the most compute-intensive step of the closed loop so ensuring the JSON-encoded scenario is correct and optimised before simulation is important in a local agentic setup. It is also important 
that all scenario are valid so if the LM hallucinates a new JSON field then Pydantic will catch the error and can provide feedback for further scenario revision.

 We use think, act, observe, update, and continue loops \cite{yao2022react} within the overall framework to generate edge cases step-by-step, as in the listing below.  
{\small
\begin{lstlisting}
[Prompt] $\rightarrow$ (LLM A: Generator) $\rightarrow$ [Draft] $\rightarrow$ {Pydantic validation} $\rightarrow$ (LLM B: Critic)
                                                                                $\downarrow$
         {Pydantic validation} $\leftarrow$ [Refined Output] $\leftarrow$ (LLM A: Update) $\leftarrow$ [Critique]  
\end{lstlisting}
}
These operate during step 1) initial scenario generation, during steps 13/14) exploit s and during step 13/14) explore scenario generation. 
This means that there are two paths through our closed loop directed by the results of the simulation evaluation metrics.  
\subsubsection{Exploitation}
Steps 1) then 13/14) exploit if the simulation identifies functional gaps during execution and create new but similar scenarios. To do this, we add 
``\texttt{Rewrite the Provided Scenario using strictly valid JSON following this interface:}'' where the interface is shown in Listing \ref{ts-spec} to step 1. This tells the LM to tweak the existing scenario.

\subsubsection{Exploration}\label{sec:exploration}
Maintaining a historical database of scenarios allows the framework to create novel scenarios during exploration by knowing what has gone before. Rather than simply generating a set of scenarios, the closed loop adapts the contextual prompt for the AI agent in step 13/14 to incorporate the scenario history so the the AI agents can adaptively target likely functional gaps.
We incorporate our historical 
data and add ``\texttt{Review the [GENERATED\_SCENARIOS\_LOG] below. Your new scenario MUST introduce a different training complexity or edge-case not covered by previous configurations.}'' to the user prompt to ask the LM to generate novel edge case scenarios.

Generally, the smaller LMs have smaller context windows as they have less working memory and can process fewer tokens (words) in their prompts. These smaller LMs can fall into ``local minima'' where they are stuck on a small set of specific archetypes and themes causing output repetition. We have deliberately created a multi-step agentic architecture using think, act, observe, update, and continue loops so we do not rely on a single prompt for scenario creation. 
Additionally, the smaller prompt length forces us to use a ``sliding window history'' implemented as a short FIFO queue of text describing scenarios (taken from the JSON generated by the LMs) as there is not enough room for the full JSON specifications or even all historical scenarios so we focus on a brief text description of recent scenarios. As new scenarios are generated then the oldest is popped and removed from the queue.
Keeping a small FIFO queue in memory is feasible and removes the overhead of running and accessing a database of past scenarios on our laptop.

The level of creativity of a LM is controlled by its temperature parameter. Using simple analyses of inspecting sets of 8 scenarios generated using different temperature values, we found that a value of temperature = 0.7 provided the best trade-off between creativity and sticking to the rules. We note that increasing the temperature and hence creativity too far can lead to the introduction of LM hallucinations.

Furthermore, using a different LM for generation, Qwen, with Nemotron for reflection aims to reduce the ``confirmation bias'' we observed when we used a single LM, Qwen, to self-critique. Minimising confirmation bias is vital for safety assurance as it reduces system robustness. This difference also provides a strict separation of LM persona, allows us to use strict rubric and engineering requirements for the different contextual prompts to the different LMs and allows us to choose models specifically for their skills such as coding models for JSON generation or reasoning models for reflection. 

\subsection{Unity-3D Simulation Environment}

Our framework is deliberately agnostic of the simulator used in step 10 (Figure \ref{fig:part2}) as we interface to the simulator using structured JSON specified by a schema. Such JSON can be ingested and produced by many programming languages and many simulators. In this paper, we use Unity-3D's ML-agents framework \cite{juliani2018unity} to test the scenarios.

ML-agents uses Unity-3D's C\# development framework as a front-end and middleware, and to interface to a Python API backend for AI training and simulation \cite{unity-ml-agents}.  In this paper, we adapt the ML-agents \textbf{Sorter} simulation and revise and augment the approach we used in \cite{hodge2021deep} by changing the task to be multistage (visiting multiple tiles in order while avoiding obstacles), improving the simulation environment so the robotic-agent can turn through 0-359$^o$ (rather than being limited to moving in the 4 cardinal directions), accelerate and decelerate, we added a LiDAR to the agent and use these sensor data during RL navigation, we enabled scenario specification in JSON format and added evaluation metric output in JSON format.

\begin{figure}[t]
\centering
\includegraphics[width=10cm]{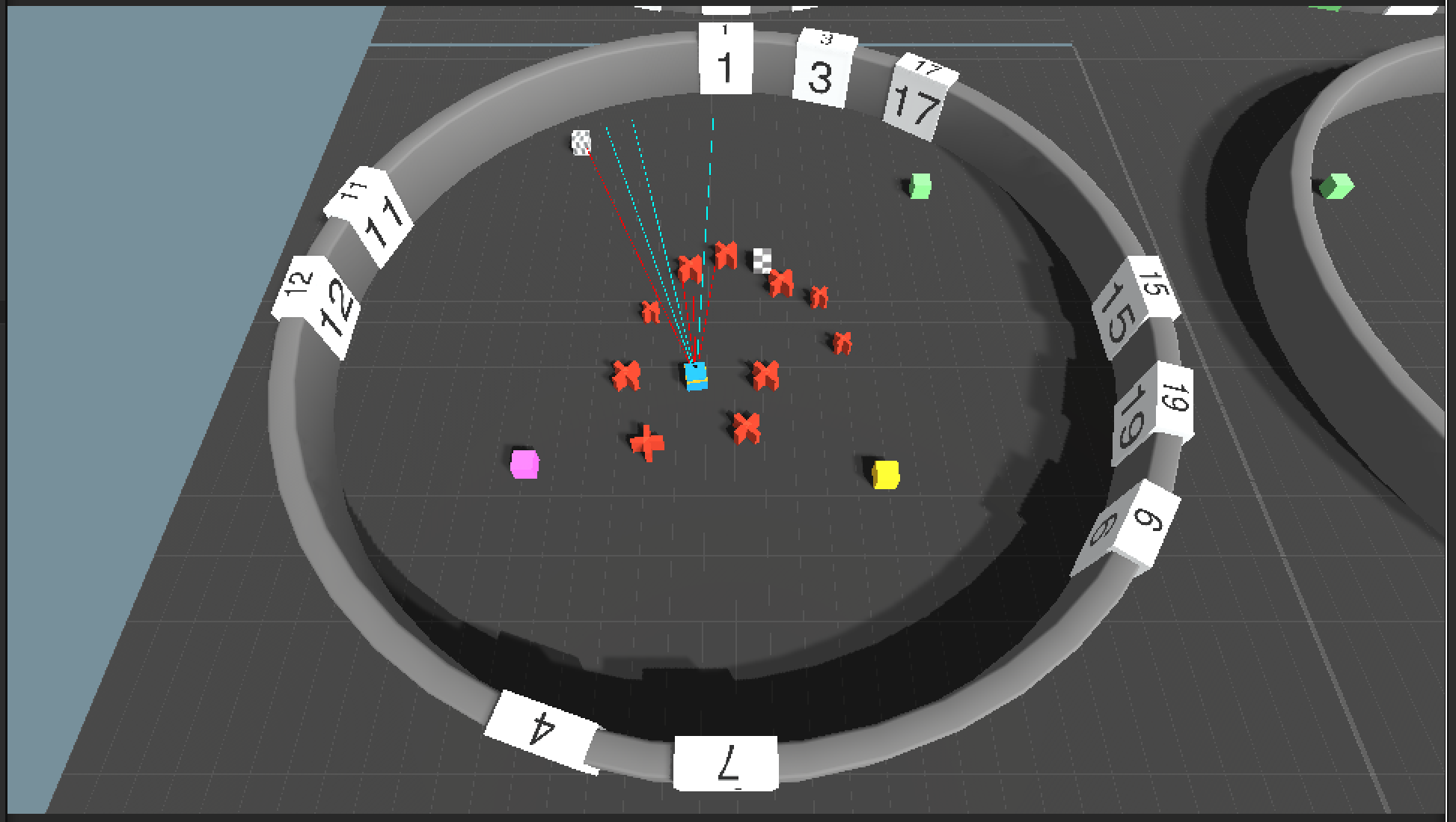}
\caption{An example scenario laid out in the Sorter arena in Unity-3D. The blue cube in the centre is the robotic agent. The red crosses are random obstacles and the coloured cubes are fixed obstacles.}\label{fig:arena}
\end{figure}

The simulation is shown in Figure \ref{fig:arena} and is a circular room with numbered tiles on the perimeter and obstacles placed randomly across the floor.  Each robotic-agent is linked to a brain (see section \ref{sec:Brains}) where we describe how  we trained the robotic-agent to navigate prior to edge case testing. \textbf{This is our system under test where we need to demonstrate that the robotic-agent's navigation is safe enough}. The robotic-agent must learn to visit the tiles in numerical value order while avoiding the obstacles. There are 5 fixed obstacles of size 1.0 (see section \ref{sec:Brains}) and each JSON-encoded scenario specifies 1-10 random obstacles of random size [1.0,3.0] and 1-20 randomly numbered and randomly placed tiles. 
We can formalise this scenario mathematically. 

Let the parameter space for all scenarios be $\Theta_{JSON}$. Each generated scenario $\mathbf{\theta_{JSON}} \in \Theta_{JSON}$ is defined by a set of obstacle positions $\mathbf{p}^o_i = (x_i, y_i, z_i)$, orientations $\mathbf{r}^o_i$, and dimensions $\mathbf{d}^o_i$ and a set of tiles with values $\mathbf{v}^t_i \in \mathbb{N} \cap [1,20])$ and associated positions $\mathbf{p}^t_i \in \mathbb{N} \cap [0,19])$. 
Thus, the JSON schema is defined by Equation \ref{eq:JSON}:
\begin{equation}
\mathbf{\theta_{JSON}} = \{(\mathbf{p}^o_1, \mathbf{r}^o_1, \mathbf{d}^o_1), (\mathbf{p}^o_2, \mathbf{r}^o_2, \mathbf{d}^o_2), \dots, (\mathbf{p}^o_m, \mathbf{r}^o_m, \mathbf{d}^o_m),(v^t_1,p^t_1),(v^t_2,p^t_2),...,(v^t_n,p^t_n)\}\label{eq:JSON}
\end{equation}
The agent navigates using LiDAR (a 7 beam ray perception sensor) which realistically  represents navigation in the real-world where the robotic-agent can only sense its local environment and may not see obstacles ahead due to occlusions. The agent receives a small penalty for each movement (-2/5000), an additional penalty (-1/5000) for reversing (to stop it travelling backwards as much as possible as it has no rear-facing LiDAR). These values were established using simple trial and error analyses of value ranges to determine the value that caused the robotic-agent to behave as desired: i.e., travel towards tiles avoiding obstacles and minimising osculation and back-tracking. The agent receives a positive reward (+1) for visiting the correct next tile, and a negative reward (-1) for either visiting the wrong tile or colliding with an obstacle. Each episode terminates if the robotic-agent reaches the highest value tile (SUCCESS), after 5000 steps (TIMES OUT), or if the robotic-agent visits the wrong tile (WRONG TILE).

\subsubsection{Evals}
In section \ref{sec:EvaluationMetrics}, we discuss the set of metrics output by the Unity-3D simulation which we use  to evaluate the quality of scenario generation.  The agentic workflow uses a subset of these as evals, as listed in Listing \ref{listing:agentic-evals} which determine if the generated scenario should be exploited or explored as shown in Listing \ref{listing:agentic-evals}. These metrics were chosen using a small experimental study to identify scenarios that reveal functional gaps and should be exploited, i.e., more similar scenarios should be generated. Otherwise, a novel set of scenarios should be generated using the historical database as guidance.
\begin{lstlisting}[language=Python, caption={Agentic AI decision process for explore or exploit.}, label={listing:agentic-evals} ]
if ( (successScore <= 45) or (crash_count > 24) 
      or (avg_ratio_tiles_visited < 0.75) ):
    exploit = True
else:
    exploit = False # and explore = True
}
\end{lstlisting}

\section{Safety Requirement to JSON Schema Mapping}\label{sec:formal}
\begin{figure}[t]
\centering
\includegraphics[width=10cm]{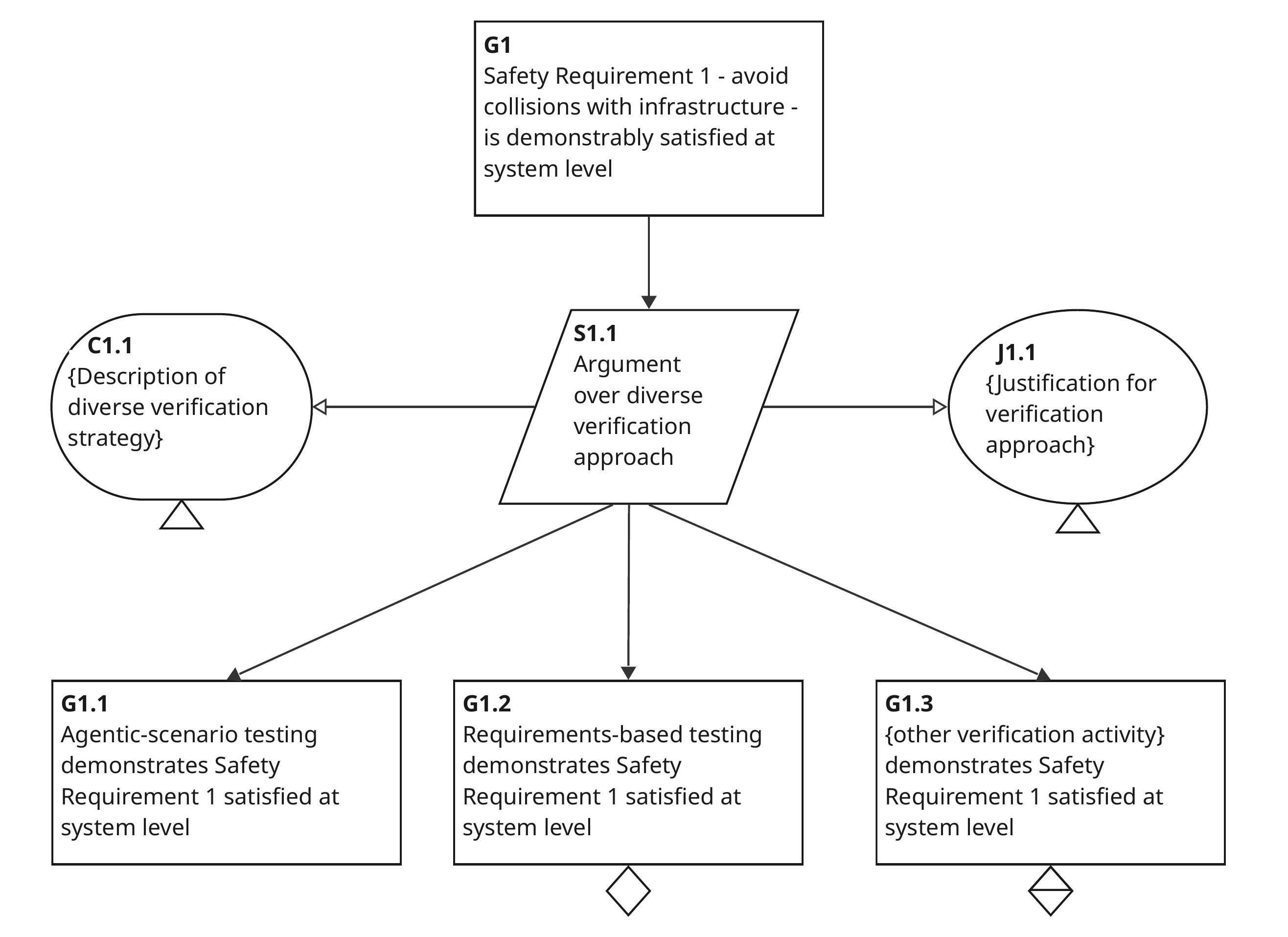}
\caption{ High-level goal structuring notation for the robotic-agent navigating the circular arena. From the claim in G1, we must demonstrate that the risk of colliding with obstacles is as low as reasonably practicable.  The GSN decomposes through levels includingt the GSN shown in Figure \ref{fig:gsnLower} }\label{fig:gsnTop}
\end{figure}

We need to take a hierarchical view when safety assuring autonomous systems by first considering system-level (product-based) safety assurance. We then work through levels of abstraction to assure the components of the autonomous system, such as the navigation process, while considering the system level requirements and goals.  ML failures lead to software failures which, in turn, leads to a system level failure and a hazard~\cite{rtca2012do178c}. Therefore, we need to argue that the navigation component meets its particular safety requirements and provide verification and validation of the associated system-level requirements. Using this hierarchical approach the claim of meeting each safety claim (goal) will be supported by other testing and argumentation. To capture the form of this argument, we use goal structuring notation (GSN) \cite{GSNstandard2021} which is widely used in academia and industry to systematically capture safety assurance argumentation. Our top claim is G1 ``\textit{avoid collisions with infrastructure - is demonstrably satisfied at system level}'' to meet safety requirement 1: ``\textit{The RL-based navigation operates the robotic-agent within a defined safety limit: it ensures zero collisions with infrastructure during navigation cycles}''. Figure \ref{fig:gsnTop} shows G1 for the autonomous system's  navigation decomposed to lower level goals G1.1, G1.2, G1.3 for the verification to demonstrate that the autonomous system's navigation component meets the claim in G1. This approach follows the methodology of SACE, specifically SACE stage 8 \cite{hawkins2026guidancesafetyassuranceautonomous}.

\begin{figure}[t]
\centering
\includegraphics[height=13cm]{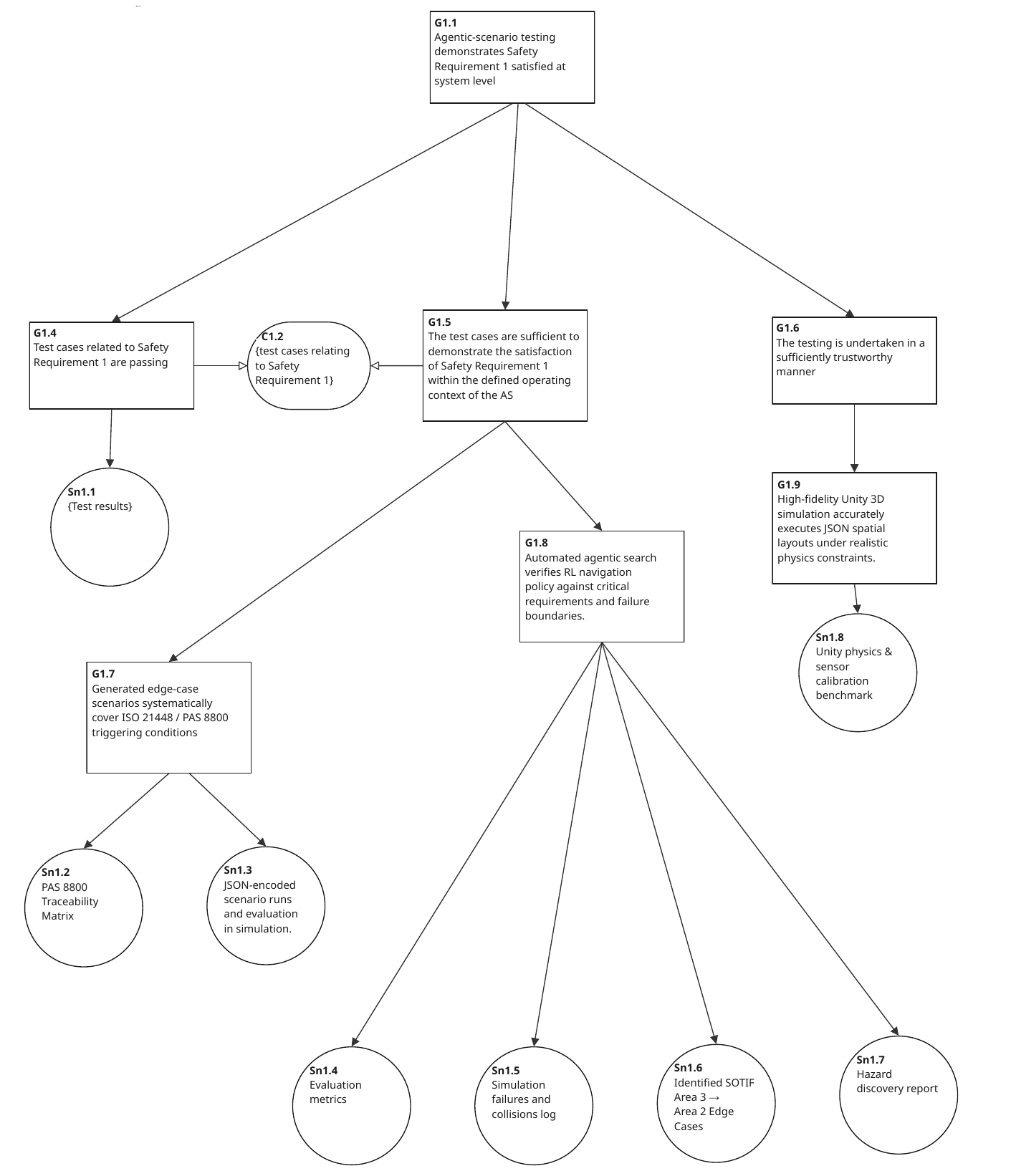}
\caption{ Lower level goal structuring notation fragment for the robotic-agent's navigation. This paper focuses on generating the evidence for Sn1.4 - Sn1.7. After generating our edge case scenarios and analysing the evaluation metrics and simulation logs, we can instantiate the artefacts for Sn1.4 \{\textit{eval metrics}\}, Sn1.5 \{\textit{simulation failures \& collisions log ($E_{sim}$)}\}, Sn1.6 \{\textit{identified SOTIF area 3 $\rightarrow$ area 2 edge cases}\} \& Sn1.7 \{\textit{hazard discovery report}\}}\label{fig:gsnLower}
\end{figure}

Our lower level safety argument is presented as GSN in Figure \ref{fig:gsnLower}.  Figure \ref{fig:gsnLower} captures the process for verification of the robot's navigation (RL algorithm) while considering the system-level requirements. The solutions (evidence requirements) for the component-level sub-goals (Sn1.1 -Sn1.8) show the evidence needed to demonstrate compliance with the claims flowing from top level G1 down to G1.1 - G1.3 and then down further to G1.4 - G1.9 for generating and evaluating scenarios specified as JSON. 

This means that overall we have two requirement levels.
\begin{enumerate}
    \item For component requirements, we need to justify that we have covered the space of scenarios that are relevant to that requirement. The aim of our edge case generation framework is to uncover unknown areas not found by other testing approaches.
    \item For the overall requirement set, we need to argue that we have covered the ODD including considering unsafe and unknown scenarios and identified any functional gaps and the triggering conditions for them, to meet ISO 21448 Clauses 7, 9, 10 and 11 and PAS 8800 stress-testing using scenario generation.
\end{enumerate}

This GSN, and the activities specified, directs the next safety assurance activities to undertake including whether to train the navigation algorithm further to fix functional gaps and what run-time assurance is required such as safety monitors to mitigate functional gaps. These would be specified at appropriate levels of abstraction and shown as parts of the overall GSN hierarchy that describes the safety assurance case.

We detail our JSON schema specification next and then show in subsequent sections how we weave together the GSN, safety assurance processes and JSON specifications for spatial layout and evaluation metrics to produce assurance evidence. 
\subsection{JSON Schema Specification}\label{sec:JSONspec}
The JSON-encoded scenario specifies the spatial layout as in equation \ref{eq:JSON} coupled with a name and text description and the start location as shown in Listing \ref{json-scenario-spec} in the Appendix. This aims to be agnostic of the LMs, simulation, task and domain.  Here we use the JSON to specify spatial layouts but JSON can specify scenery, weather or dynamic objects and other ODD factors. An LM such as qwen3-coder:30b will have been trained on more software code than JSON schema files. Listing \ref{ts-spec} converts our JSON scenario schema to a Typescript interface (schema) which we use in our contextual prompt for JSON-encoded scenario generation. We found smaller LMs were more effective and produced more concise and readable JSON when the prompt includes the schema specified as Typescript but requests JSON output. Once the generate-critique-refine loop has completed, we verify the JSON using the Pydantic specification in Listing \ref{pydantic-spec}. 

The simulation also outputs the values for the evaluation metrics in JSON so we can again use Pydantic to parse and verify this JSON. Our Pydantic specification for parsing this JSON is shown in Listing \ref{listing:evalJSON}. The Pydantic specification can easily be varied for different sets of evaluation metrics. as the sets will vary per environment, system and task.

\subsection{Constraint Validation}\label{sec:constraintVal}

Our domain ontology represents the ODD and comprises the environment description, system description and system requirements. Each section of the domain ontology  along with the task description is a specific ``knowledge base'' (KB) providing guidance for the LM with respect to the spatial layout, physical laws that must be obeyed, the data schema, and logical goals of Unity-3D ML-Agents environment and what to generate.  The knowledge bases combine in our contextual (knowledge-based) prompts passed to the LM, similar to \cite{bae2026wargames}.  The KBs are encoded as YAML markup which can define entities and their relationships similar to a database but in a human readable format. This YAML specification allows us to specify requirements as non-negotiable constraints on the system and environment. It provides contextual alignment and reduces hallucination by focusing on the specifics of the task and helps to ensure valid JSON outputs. The environment description is given in Listing \ref{listing:envDescription}, system description in Listing \ref{listing:sysDescription} \& system requirements in Listing \ref{listing:sysReqs}) along with the task description in Listing \ref{listing:taskDescription}. The KBs are included in the prompts for 1) initial JSON-encoded scenarios, 4) critique, 6) revise using critique, 13/14) exploit using successful JSON-encoded scenario and 13/14) explore by creating novel JSON-encoded scenarios using history. 
Combining The KBs in contextual prompts aims to ensure that the agentic framework generates JSON-encoded scenarios that are valid w.r.t. the ODD and JSON schema. 

To determine the usefulness of the KBs in the LM prompts, we ran a mini ablation study where we removed the \{\textit{env description}\} and then the \{\textit{system description}\} in turn to determine the effects. The scenarios generated no longer obey the rules. For example, the \{\textit{env description}\} specifies that obstacles are scaled in the range [1.0,3.0] but removing that produced larger obstacles such as scale = 4.0.
The \{\textit{system description}\} specifies that ``\texttt{Ensure that no obstacle is at x=0.0, z=0.0 or overlaps x=0.0, z=0.0 to avoid the agent's start place}'' but the LM places obstacles at (x=0.0,z=0.0) if the \{\textit{system description}\} is omitted.

Despite these KBs providing non-negotiable constraints, LMs are probabilistic and susceptible to ignoring specifications. To mitigate this, we exploit our reflection (steps 4-6) to both critique/validate then refine the JSON-encoded scenario against the specifications in the knowledge bases, as well as running syntactic validation against the schema in step 8. 

\subsection{SOTIF Trigger Mapping}\label{sec:sotifTriggerMapping}
ISO 21448: Clause 7 mandates the identification and analysis of functional gaps and their triggering conditions. By analysing the rich text descriptions generated by LMs, we are able to find the specific environmental, geometric, or infrastructure states that cause functional gaps. From these gaps, we can make functional modifications to mitigate any unacceptably high risks to satisfy SOTIF clause 8.  The methodology we have detailed so far feeds into providing evidence that clauses 9, 10 and 11 (see section \ref{sec:Introduction}) are satisfied. We now detail how our approach can satisfy clause 7.

For clause 7 we need to identify where:
$$\text{JSON-encoded Scenario } (\mathbf{\theta}_{JSON}) \xrightarrow{\text{in Unity-3D}} \text{Triggering Condition } (TC) \xrightarrow{\text{Exposes}} \text{Functional Gap } (FG)$$ The functional gaps are found by processing the JSON-formatted evaluation metrics output by Unity-3D and mapping these to the text descriptions in $\theta_{JSON}$ (see equation \ref{eq:JSON}). 

A scenario $\mathbf{\theta_{JSON}}$ is a SOTIF triggering condition ($TC$) if, at any point in time $t$ from the start of the simulation until the scenario completes or times out $[0,T]$, executing $\mathbf{\theta_{JSON}}$ exposes a functional gap ($FG$) that violates a safety constraint for the autonomous system, i.e., $g(\mathbf{AS}) < 0$
$$TC(\mathbf{\theta_{JSON}}) = \begin{cases} 1 & \text{if } \exists t \in [0, T] \text{ s.t. } g(\mathbf{AS}(t)) < 0 \quad (\text{Functional gap exposed}) \\ 0 & \text{otherwise } (\text{Area 1 / Known Safe}) \end{cases}$$
Where $g(\mathbf{AS})$ represents required safety guarantees on the autonomous system such as minimum obstacle clearance,
maximum allowable incorrect tile visits, or path completion time bounds.

To demonstrate that we have met our required safety guarantees $g(\mathbf{AS})$, we gather evidence by:
\begin{enumerate}
    \item \textbf{Determining triggering conditions}: $\theta_{JSON}$ and Unity-3D specify the operating domain using a discretized symbolic coordinate space, rather than a continuous high-dimensional rendering space. $\theta_{JSON}$ contains a semantic description of the scenario.  This means we can map high-level ISO 21448 Triggering Condition semantics (e.g., occlusion, clutter, narrow passage) directly into executable spatial layouts. The JSON output of the Unity-3D execution allows us to identify the SOTIF $TC(\mathbf{\theta_{JSON}})$ that has triggered.
    \item \textbf{Automated discovery of area 3 scenarios}: when the evaluation metrics output by Unity-3D reveal that a JSON-encoded scenario layout $\theta_{JSON}$ caused the navigation stack to violate safety boundaries, an ISO 21448 area 3 (unknown/unsafe) scenario is flagged. The safety engineers can then convert this to an area 2 (known/unsafe) and develop suitable mitigations to move $\theta_{JSON}$ to area 1 and to demonstrate that we have ensured the risk is at an acceptable level. The JSON-encoded scenario $\theta_{JSON}$ in equation \ref{eq:JSON} and its text description form a reproducible artifact and explanation of the specific triggering condition.
    \begin{enumerate}
        \item \textbf{Semantic safety}: Furthermore, AI agents are able to understand human-like safety rules and requirements (e.g., ``don't go near tables'') and, by including these rules and requirements in the prompts, the agents can generate scenarios involving those specific rules. The use of simulation enables visual inspection of the running scenarios to validate these rules.
    \end{enumerate}
    \item \textbf{Traceability to safety cases}: ISO/PAS 8800 mandates that scenario generation tools stress-test the autonomous system navigation and it recommends scenario-based testing and the usage of synthetic data environments (simulation). The pair $(\mathbf{\theta}_{JSON}, \text{Eval\_Metrics})$ provides direct empirical evidence Sn1.3 and Sn1.4 for GSN towards claims G1.7 \& G1.8 in Figure \ref{fig:gsnLower} following PAS 8800 using a traceability matrix Sn1.2, arguing that functional gaps \{FG\} under specific triggering conditions \{TC\} have been systematically tested and logged.
\end{enumerate}

\section{Experimental Setup \& Benchmarking}\label{sec:experiments}
Our use case is the simulation setup described in Listing \ref{listing:Unity-setup-YAML} and shown in Figure \ref{fig:arena} but our agentic scenario generation framework is agnostic of the simulation and use case. Next we describe how we trained the robotic-agent to be tested in our framework.

\subsection{Simulation Training}
Our lightweight JSON schemas (for specifying scenarios and capturing evaluation results) are agnostic of the simulator and system. The system might have a complex control architecture and be distributed with redundancy or voting-like systems and a complex hardware architecture to match. The robotic-agent and controller in this paper are simpler but the framework and process can be adapted to more complex environments, systems and tasks.

The robotic controller under test has a ``brain'' \cite{hodge2021deep} that determines the optimal action to take given the current state of the system (robot), the environment and the task and its goal, as described next. The brain is what we are stress-testing here with our edge-case scenarios to pinpoint functional gaps.
\subsubsection{Brains}\label{sec:Brains} 
The brain provides the decision logic and determines the best action to take at each state. As in \cite{hodge2021deep}, we use deep reinforcement learning (deep RL) which is an experience-driven learning approach that learns by generating rewards and penalties as the robot navigates. We use this to demonstrate how to ensure that an adaptive trial and error navigation algorithm is thoroughly tested using scenario-based testing and how we uncover any gaps in the RL policy where the robotic-agent has not learned the correct actions to take so is not acceptably safe.

Formally, RL is a Markov decision process (MDP), comprising: 
\begin{itemize}
\item A finite set of states $S$, plus a distribution of starting states $p(s_0)$. There may be a terminal state, $s_T$. 
\item A set of actions $A$ available in each state. 
\item A policy $\pi_\theta (s_{t+1}|s_t, a_t)$ that maps state/action pairs at time $t$ onto a distribution of states at time $t + 1$ using parameter set $\theta$. 
\item An instantaneous reward function $R(s_t, a_t, s_{t+1})$ associated with each transition to $s_{t+1}$.
\item A discount factor $\gamma \in [0, 1]$, which is the current value of future rewards. It quantifies the difference in importance between immediate rewards and future rewards (lower values place more emphasis on immediate rewards). 
\end{itemize}
We treat our operational domain as a partially observable MDP or POMDP as the LiDAR only detects obstacles that intercept its beams so obstacles ahead may be obscured and unknown. In a POMDP, the agent receives an observation $o_t \in \Omega$, where its distribution $p(o_{t}|s_{t}, a_{t-1})$ depends on the previous action $a_{t-1}$ and the current state $s_{t}$. Here the observation is described by a mapping in a state-space-model that depends on the current state. 

RL learns a parameterised policy function $\pi$ which gives the probability of taking action $a_{t}$ given the current state $s_{t}$ according to the deep RL model with weights parameters $\theta$. Thus, $\pi_\theta (a_t|s_t) = P [A_t = a_t | S_t = s_t]$. A policy fully defines the behaviour of the robotic-agent given the current state, and that action, when executed, generates a reward $r_t$. The aim of RL is to identify the optimal policy, $\pi^{*}$, which maximises the reward over all states (i.e. maximises the expected reward value $E$):  $\pi^{*} = argmax_\pi \, E[R_t|\pi]$. As in our previous paper,\cite{hodge2021deep}, we use proximal policy optimisation (PPO) \cite{schulman2017proximal} which computes an update at each time step $t$ that minimizes the cost function while ensuring the deviation from the previous policy is relatively small. PPO prevents large, destructive policy changes, making it highly efficient for training autonomous systems.
In \cite{hodge2021deep}, we developed incremental curriculum learning for training our robotic-agent's using PPO which we describe next.

\subsubsection{Incremental Curriculum Learning}
Curriculum learning is used 
for autonomous system development as it allows the RL to begin learning in simulation and then transition and continue learning in the real-world. Curriculum learning begins with a simple task and gradually increases the task complexity until it reaches the required operational domain. We began with up to 2 tiles with values 1 \& 2 placed randomly, 5 fixed obstacles in different positions per arena to provide continuity during training and evaluation and 5 randomly placed obstacles all with size 1.0. We increment the number of tiles by 2 per lesson (up to 20). Curriculum learning uses transfer learning to transfer the current policy $\pi_\theta (a_t|s_t)$ from each lesson to the next. Our incremental curriculum learning approach \cite{hodge2021deep}  trains the RL brain during a lesson until a metric baseline is exceeded; i.e., until we are satisfied that the PPO has learned that lesson sufficiently. In contrast, traditional curriculum learning pre-specifies the number of epochs for each lesson in advance which may not be sufficient or may cause over-training. We analysed a number of metrics as noted in our previous paper \cite{hodge2021deep} and found that mean final reward generated the best model (compared to using loss, entropy or steps taken). 

\subsubsection{Training}
Unity-3D ML-Agents allows parallel learning and we were able to run eight instances of the domain in parallel (as in Figure \ref{fig:8-arenas}) which speeds up training, allowing the environment to gather different experiences in parallel. We trained the PPO brain for 15 million training epochs (scenarios) using our incremental curriculum learning as shown in Listings \ref{PPO-setup-YAML} and \ref{curriculum-yaml} which has stopping criteria based on the reward score to determine when to cease learning each curriculum lesson. Each scenario is generated independently and randomly as a separate navigation task using Unity-3D's C\# random number generator \cite{hodge2021deep}. The 1-20 tiles and 5 moveable obstacles are arranged in the grid according to the random layout. This trains the brain for the evaluations here.

\subsection{Scenario-Based Testing}
\begin{figure}[t]
\centering
\includegraphics[width=\linewidth]{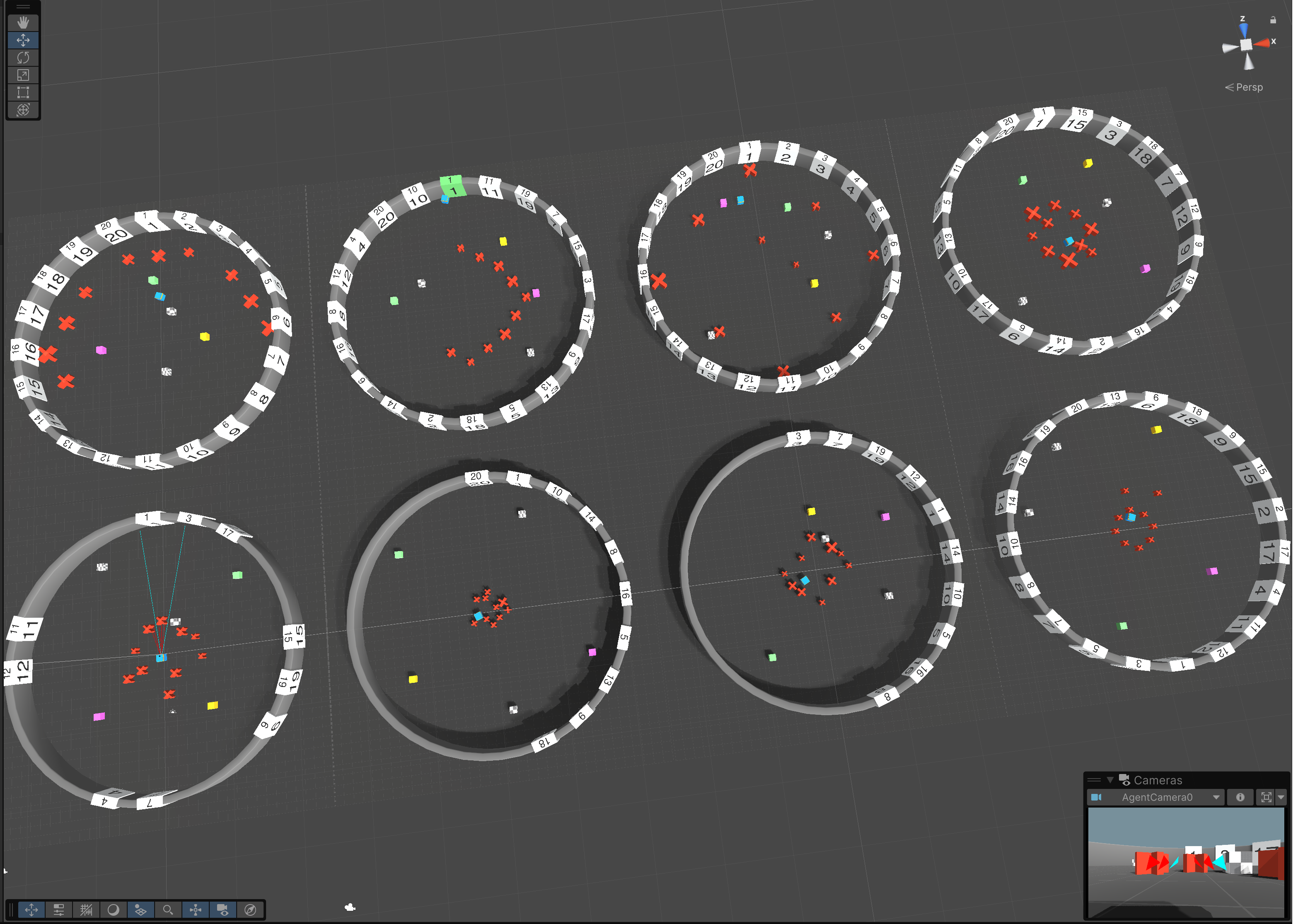}
\caption{Eight scenarios generated by the Agentic AI and laid out in the Sorter arena in Unity-3D.}\label{fig:8-arenas}
\end{figure}

To test the effectiveness of the trained RL brain w.r.t. coverage, robustness and accuracy, and to uncover functional gaps in its navigation, we analyse the agentic framework against two baseline scenario generation approaches described next. Each approach generates 512 scenarios in total in batches of 8 that follow the JSON schema in Listing \ref{json-scenario-spec}. Each batch run in parallel in Unity-3D as shown in Figure \ref{fig:8-arenas} for 50 episodes per scenario, giving 512 x 50 =25,600 episodes per approach. PPO models are non-deterministic as PPO uses a stochastic  policy designed to explore its environment by sampling actions from a probability distribution. Hence, we run 50 episodes per scenario to factor in this stochastic action selection, and potential randomness within the environment itself.

\subsubsection{Baselines for Comparison}
The two baselines are listed in Table \ref{tab:baselines} and are inspired by the baselines in our previous paper \cite{hodge2021deep}. We generate and analyse a set of metrics, described next, to provide an all round view of the scenario generation performance.
\begin{table}
	\caption{The baseline scenario generation approaches to allow us to evaluate the merits of our agentic approach.}
	\centering
\ra{1.2}
{
\small
	\begin{tabular}{p{3.4cm}p{10.6cm}}
		\toprule
		  Baseline & Description \\
		\midrule
		     \textbf{Random\_1} & Random asset placement generated using the Python random number generator with 1-20 tiles and 1-10 obstacles. Tile values are generated as unique values (used at most once). If tile or obstacle locations clash then a new location is generated.\\
          \textbf{Random\_5} & Random asset placement generated using the Python random number generator with 1-20 tiles and \textbf{5}-10 obstacles. Tile values are generated as unique values (used at most once). If tile or obstacle locations clash then a new location is generated.\\
		\bottomrule
	\end{tabular}
}
	\label{tab:baselines}
\end{table}

\subsection{Evaluation Metrics}\label{sec:EvaluationMetrics}
We developed a large set of scenario evaluation metrics which are output by the Unity-3D simulation as JSON. 
These aim to evaluate the important factors for scenario generation including those listed in \cite{deGelder2025vehicles}. Our metrics  are listed in Table \ref{tab:evalMetrics} and assess breadth of coverage, the complexity (time to solve), the diversity, the level of difficulty (hazards), and the frequency of hazards. We also include charts showing the spatial heatmap of obstacle placements, how many times each number (count) of obstacles was selected,  how many times each number (count) of tiles was selected, how many times each tile position was selected and  how many times each tile value was selected.  Success rate, timeout score and wrong tile score coupled with average reward, percentage of tiles visited and collision count measure the difficulty of scenarios as they measure functional gaps. The spatial heatmap of obstacle locations coupled with the obstacle counts, tile counts, tile positions and tile locations demonstrate coverage. Ideal path length measures the complexity and its spread measures the diversity. Average path length, action transition entropy and action distribution entropy measure the complexity, diversity and hazard exposure.

\begin{table}
	\caption{The set of evaluation metrics exported by Unity-3D for use in the agentic AI closed loop reflection analyses.}
	\centering
\ra{1.2}
{
\small
	\begin{tabular}{p{3.4cm}p{10.6cm}}
		\toprule
		  Metric & Description \\
		\midrule
		     Success score & how many times it successfully completed the task (over 50 episodes)\\
             \myrowcolour
          Timeout score & how many times the task timed out (over 50 episodes)\\
          Wrong tile score & how many times it visited the wrong tile (over 50 episodes)\\
             \myrowcolour
          Avg path length & the average number of steps taken (over 50 episodes)\\
          Ideal path length & the ideal number of steps (shortest path)\\
             \myrowcolour
          Regret  & (avgPathLength / idealPathLength)\\
          Avg \%ge tiles visited &  average percentage of tiles it successful visited (over 50 episodes)\\
             \myrowcolour
          Avg reward & average reward accumulated at the end of each episode (over 50 episodes)\\
          Action distribution entropy (unigram) & $H(A) = -\sum_{i=1}^k P(a_i)log_2P(a_i)$ Where $P(a_i)$ is the probability (frequency) of action $a_i$ occurring, and $K$ is the total number of unique actions available. We normalize this between 0.0 (always choosing the exact same action) and 1.0 (choosing all actions perfectly equally).\\
             \myrowcolour
          Action transition entropy (bigram) & measures the predictability of the next action given the current action. It is the key to identifying patterns like [Turn Left, Turn Left, Turn Left, Turn Left]. If an agent is spinning in circles, its transition entropy will collapse toward 0, because "Turn Left" is always followed by "Turn Left." \\
          Collision count &  how many obstacles the robotic-agent collided with (over 50 episodes)\\
		\bottomrule
	\end{tabular}
}
	\label{tab:evalMetrics}
\end{table}

\section{Results}\label{sec:results}
We use the evaluation metrics in Table \ref{tab:evalMetrics} coupled with the spatial layouts in the JSON-encoded scenarios as defined in Equation \ref{eq:JSON} to produce charts for the three scenario generation techniques (Random\_1, Random\_5, Agentic). These provide an all round view for assessing the utility and quality of scenario generation methods. The next section details the results which are discussed and analysed in Section \ref{sec:discussion}.

\subsection{Failure Discovery Efficiency}
\begin{figure}[ht]
\begin{minipage}[b]{.50\textwidth}
\centering
\includegraphics[width=1\textwidth]{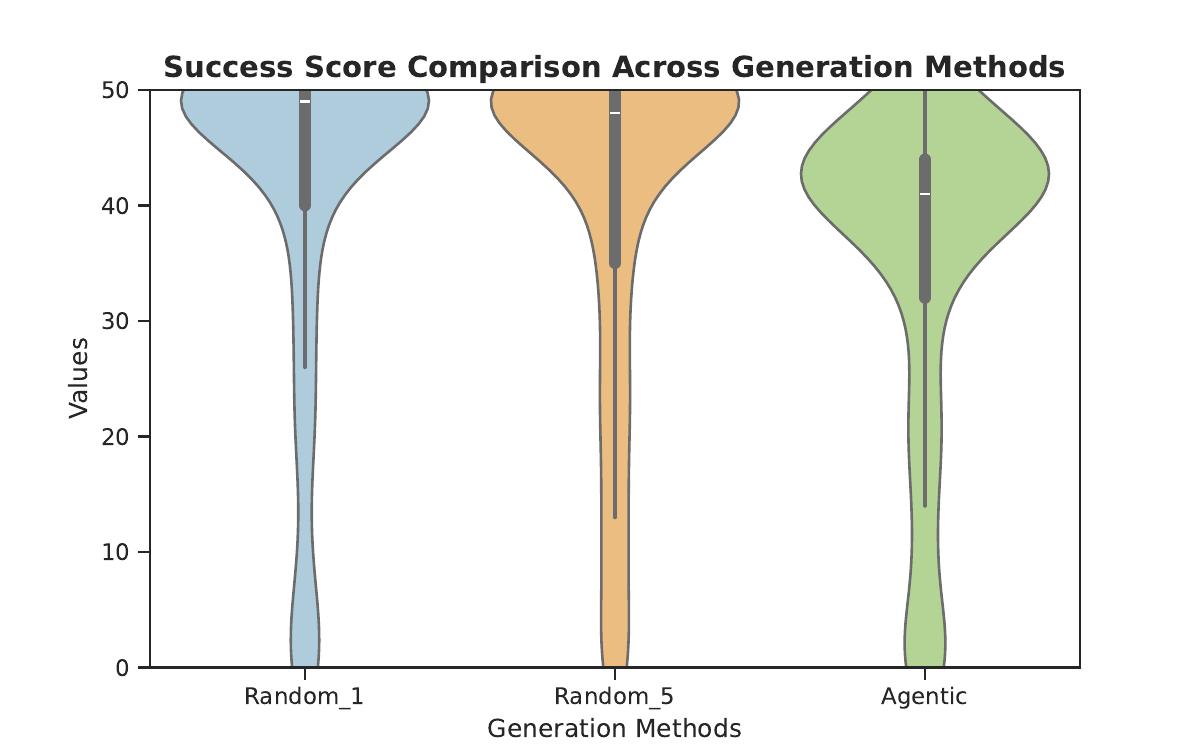}
\caption{Success ratio for each scenario generation method - how many times it completed each scenario over 50 repetitions}\label{fig:success}
\end{minipage}
\hfill
\begin{minipage}[b]{.48\textwidth}
\centering
\includegraphics[width=1\textwidth]{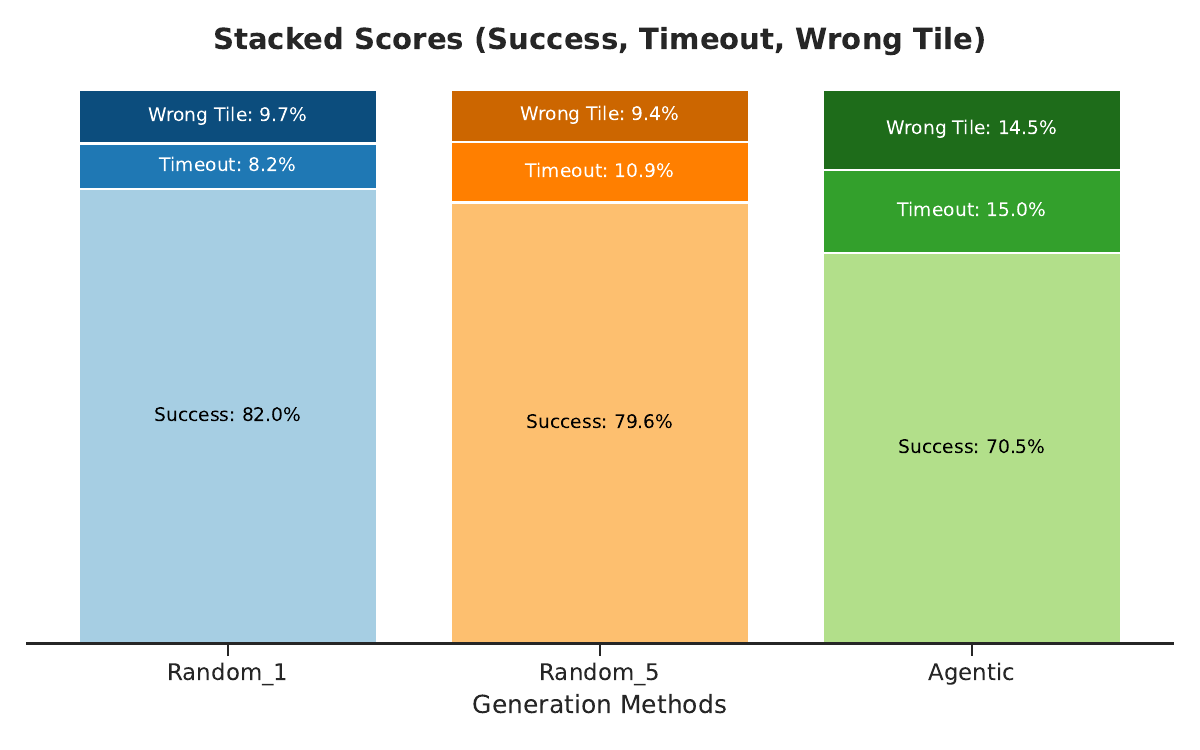}
\caption{Stacked bars comparing the ratio of success vs timeout vs wrong tile visited for each scenario generation method.}\label{fig:stackRatios}
\end{minipage}
\end{figure}

\begin{figure}[ht]
\begin{minipage}[b]{.49\textwidth}
\centering
\includegraphics[width=1\textwidth]{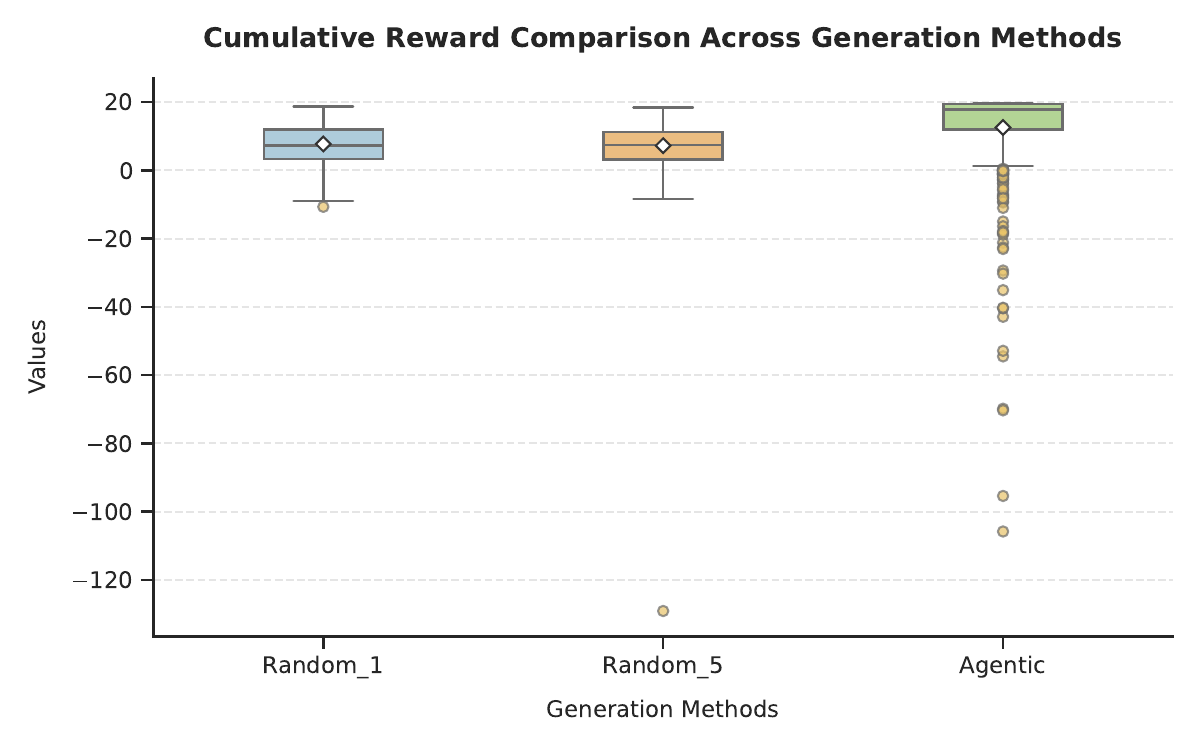}
\caption{Chart showing the average cumulative reward across all runs for each scenario generation method.}\label{fig:reward}
\end{minipage}
\hfill
\begin{minipage}[b]{.49\textwidth}
\centering
\includegraphics[width=1\textwidth]{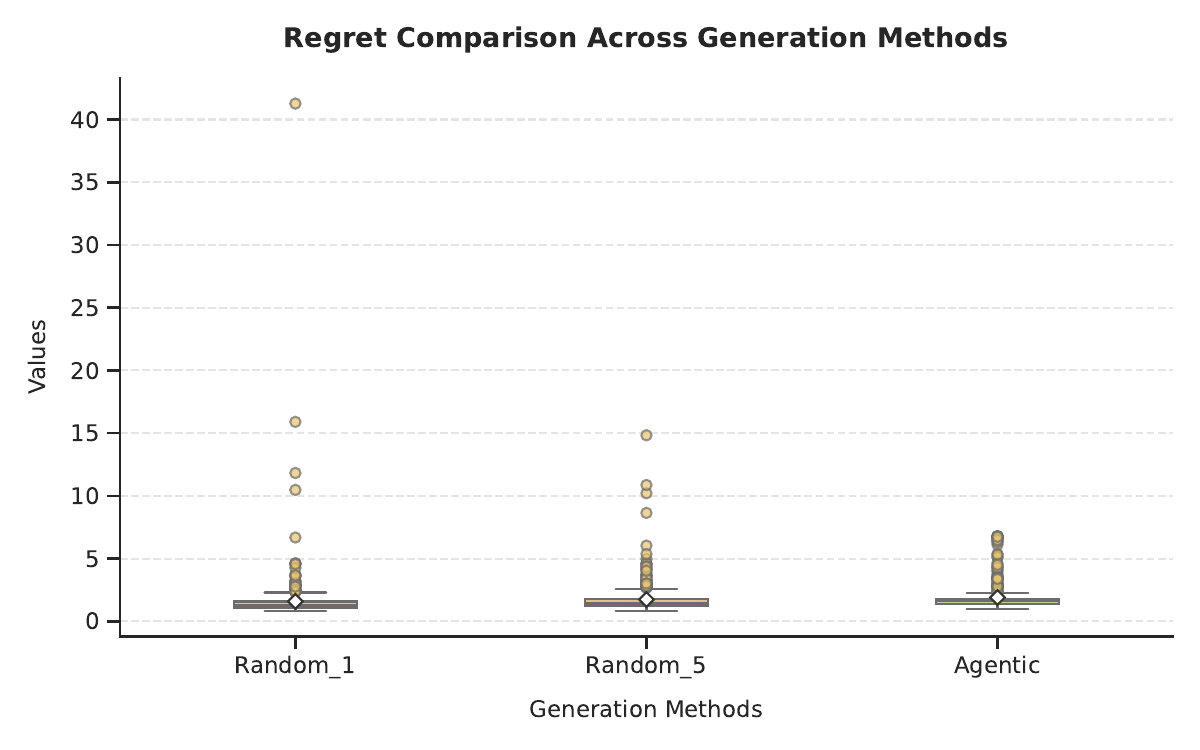}
\caption{Chart showing the regret (average path length / ideal path length) across all runs for each scenario generation method.}\label{fig:regret}
\end{minipage}
\end{figure}

\begin{figure}[ht]
\begin{minipage}[b]{.49\textwidth}
\centering
\includegraphics[width=1\textwidth]{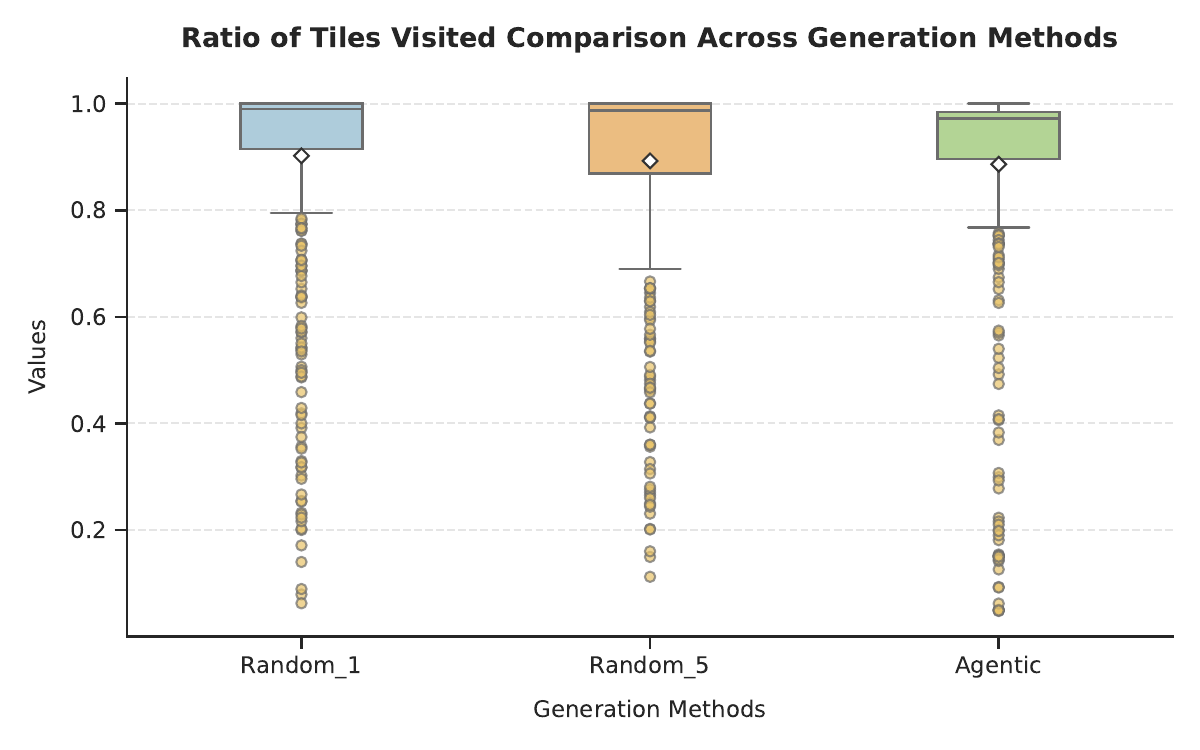}
\caption{Charts showing the percentage of all tiles successfully visited across all runs for each scenario generation method.}\label{fig:tilesVisited}
\end{minipage}
\hfill
\begin{minipage}[b]{.49\textwidth}
\centering
\includegraphics[width=1\textwidth]{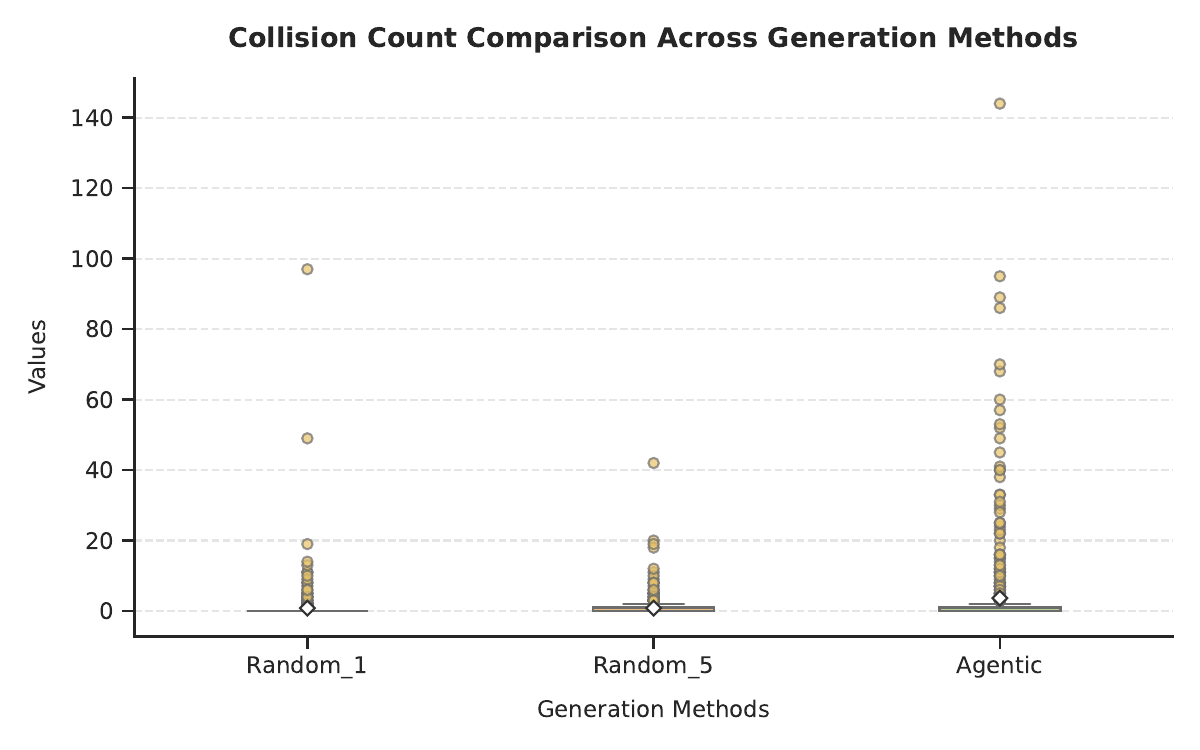}
\caption{Chart showing the collision counts per 50 repetitions across all runs for each scenario generation method.}\label{fig:collisionCDF}
\end{minipage}
\end{figure}

The charts in Figure \ref{fig:success} and \ref{fig:stackRatios} show the ratio of successful and unsuccessful scenarios, Figure \ref{fig:reward} shows the final reward chart while the chart in Figure \ref{fig:regret} shows the regret. Together these demonstrate how successful each technique is at generating critical scenarios that uncover functional gaps. Figures \ref{fig:tilesVisited} \& \ref{fig:collisionCDF} show the ratio of tiles visited and the number of obstacle collisions. These show how difficult the scenarios are to complete successfully.

\begin{figure}[t]
\centering
\includegraphics[width=8cm]{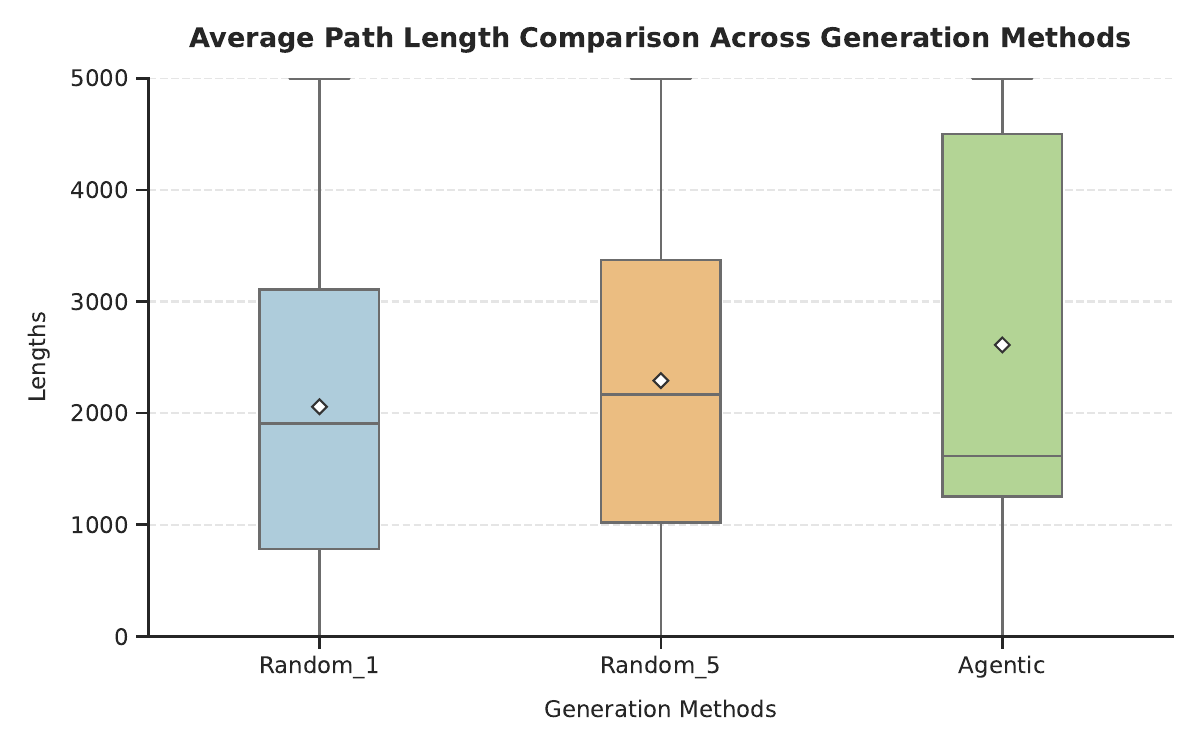}
\caption{Chart showing the average path length across all scenarios for each scenario generation method. The path length is 0 if it visited the wrong tile.}\label{fig:avgPath}
\end{figure}

\begin{figure}[ht]
\begin{minipage}[b]{.49\textwidth}
\centering
\includegraphics[width=1\textwidth]{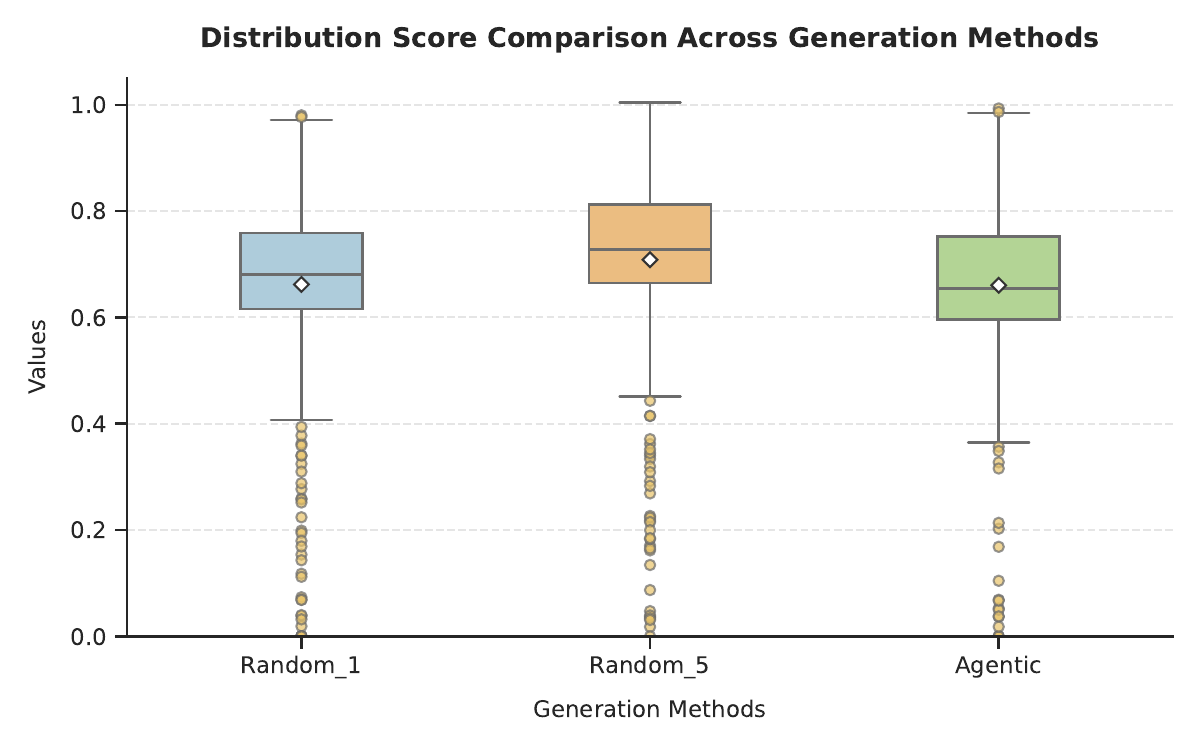}
\caption{Chart showing the action distribution entropy across all runs for each scenario generation method.}\label{fig:distribScore}
\end{minipage}
\hfill
\begin{minipage}[b]{.49\textwidth}
\centering
\includegraphics[width=1\textwidth]{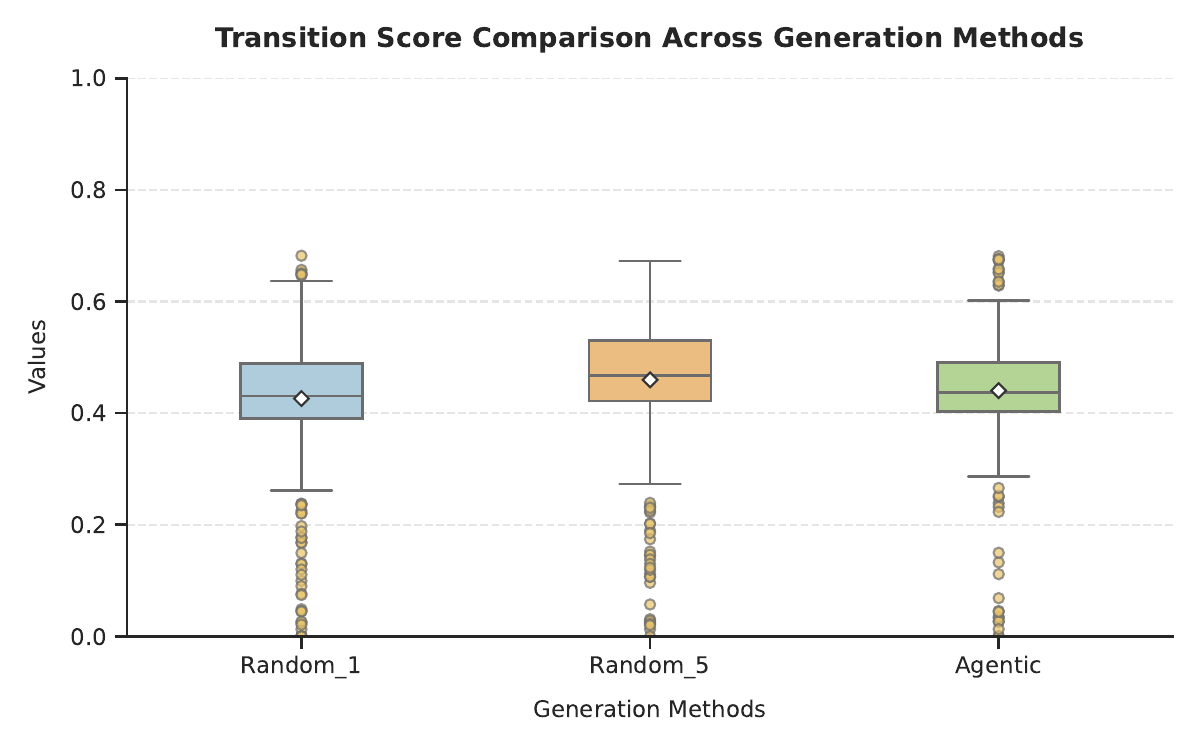}
\caption{Chart showing the action transition entropy across all runs for each scenario generation method.}\label{fig:transitScore}
\end{minipage}
\end{figure}

Figure \ref{fig:avgPath} shows the chart of average path length, Figure \ref{fig:distribScore} shows the action distribution entropy chart while Figure \ref{fig:transitScore} shows the chart of action transition entropy. Together these measure the complexity, diversity and exposure of the scenarios generated as more complex scenarios lead to longer path lengths and expose where the robot prefers certain actions or displays repetitive actions. The inter-quartile ranges show the diversity of the scenarios.

\subsubsection{Spatial Heatmaps}

\begin{figure}[t]
\centering
\begin{tabular}{ccc}
\subf{\includegraphics[width=45mm]{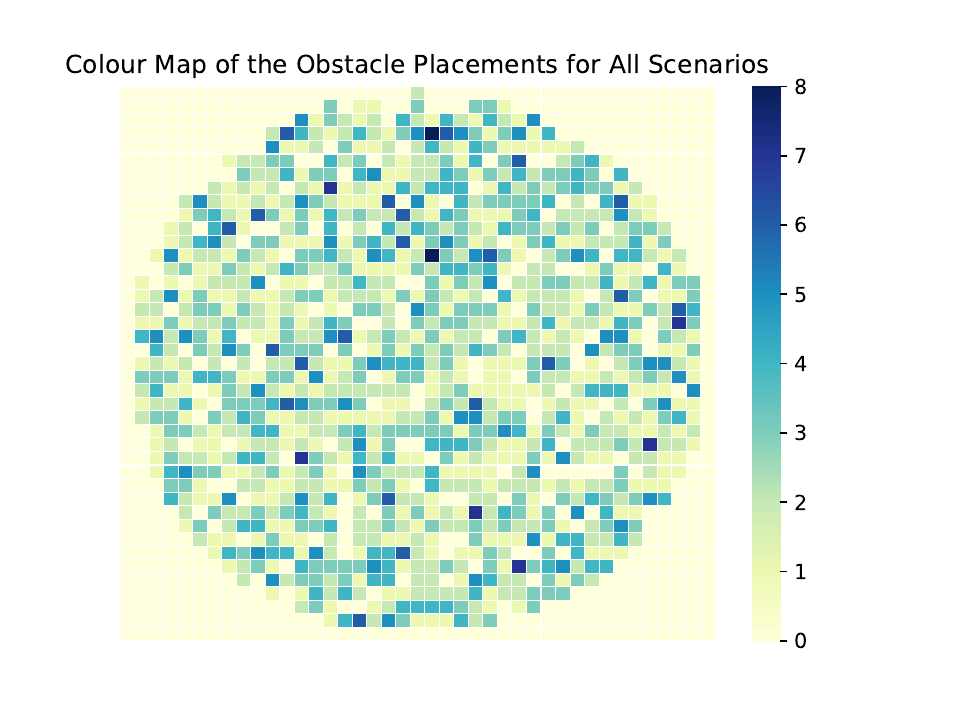}}
     {Random\_1}
&
\subf{\includegraphics[width=45mm]{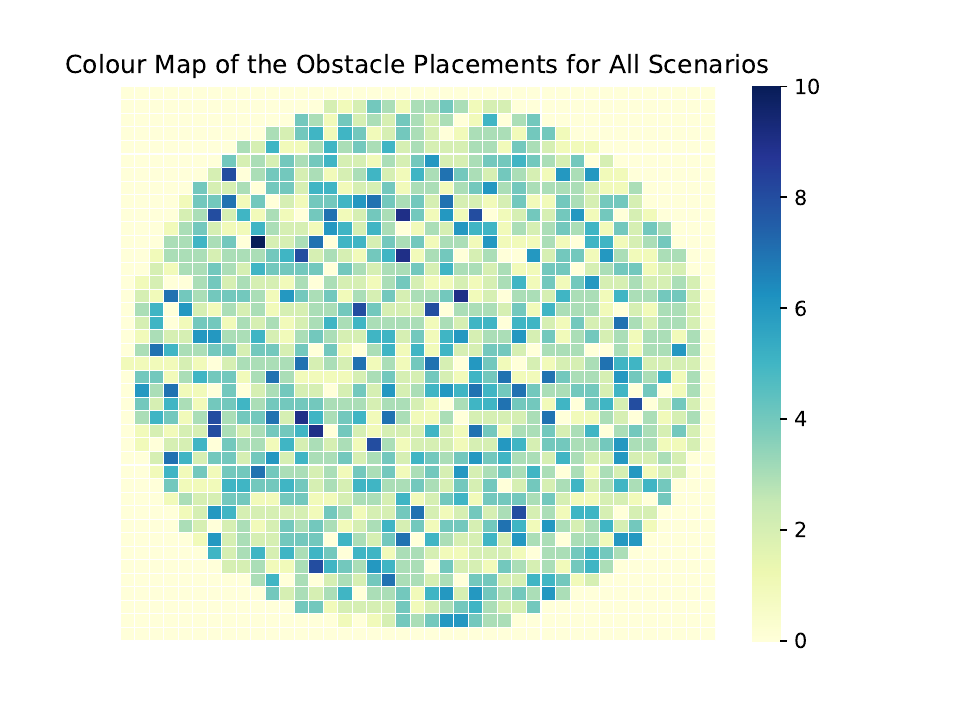}}
     {Random\_5}
&
\subf{\includegraphics[width=45mm]{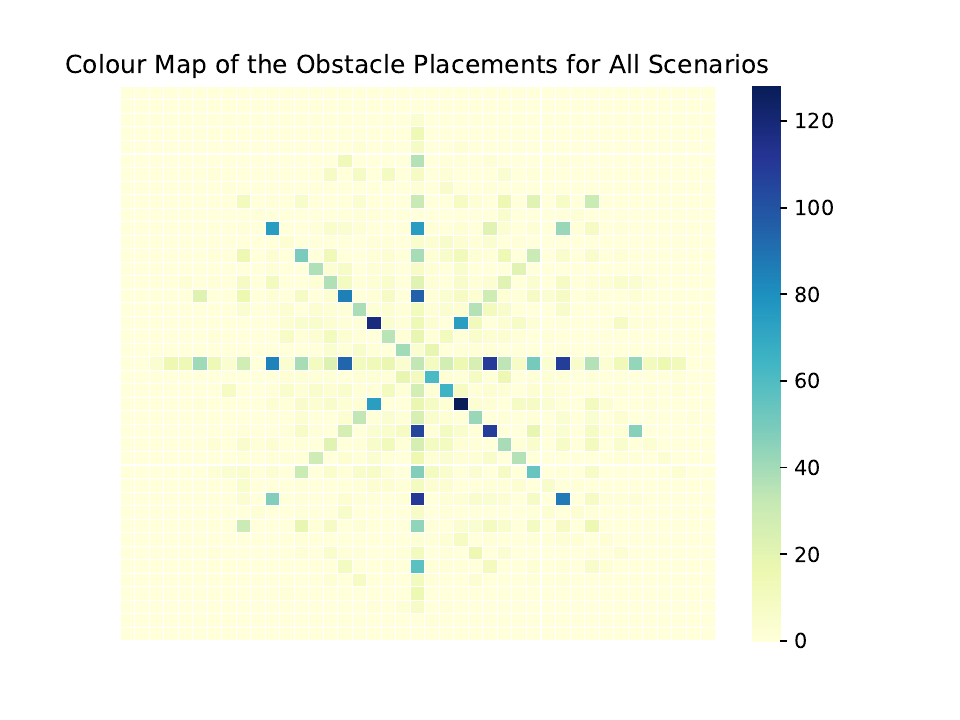}}
     {Agentic}
\\
\end{tabular}
\caption{Spatial maps of obstacle placements across all scenarios for each scenario generation method.}\label{fig:heatmap}
\end{figure}

\begin{figure}[t]
\centering
\begin{tabular}{ccc}
\subf{\includegraphics[width=45mm]{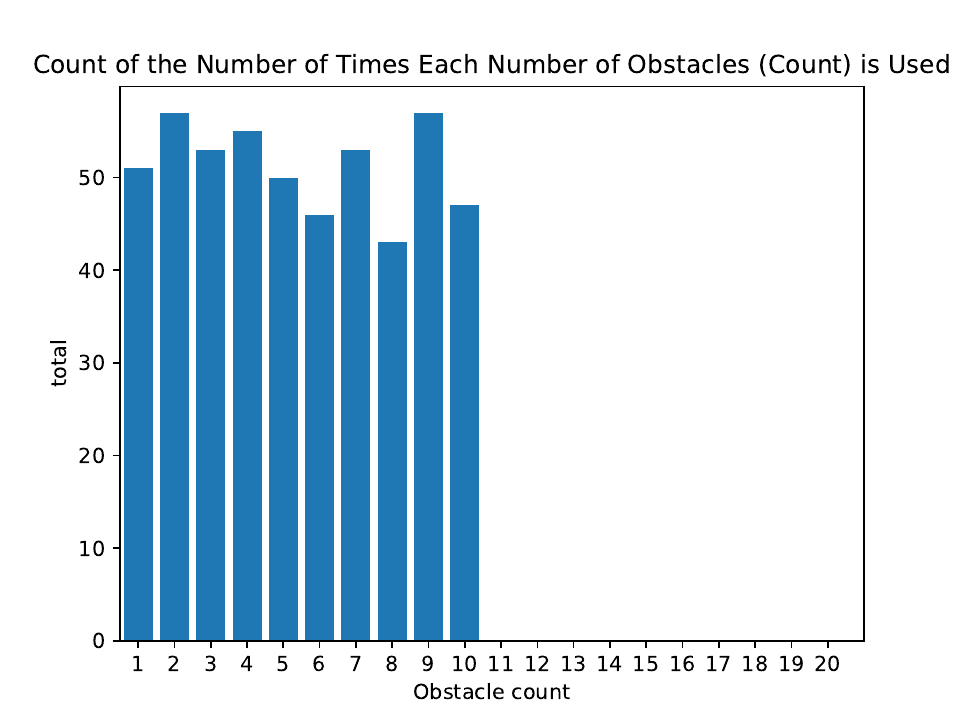}}
     {Random\_1}
&
\subf{\includegraphics[width=45mm]{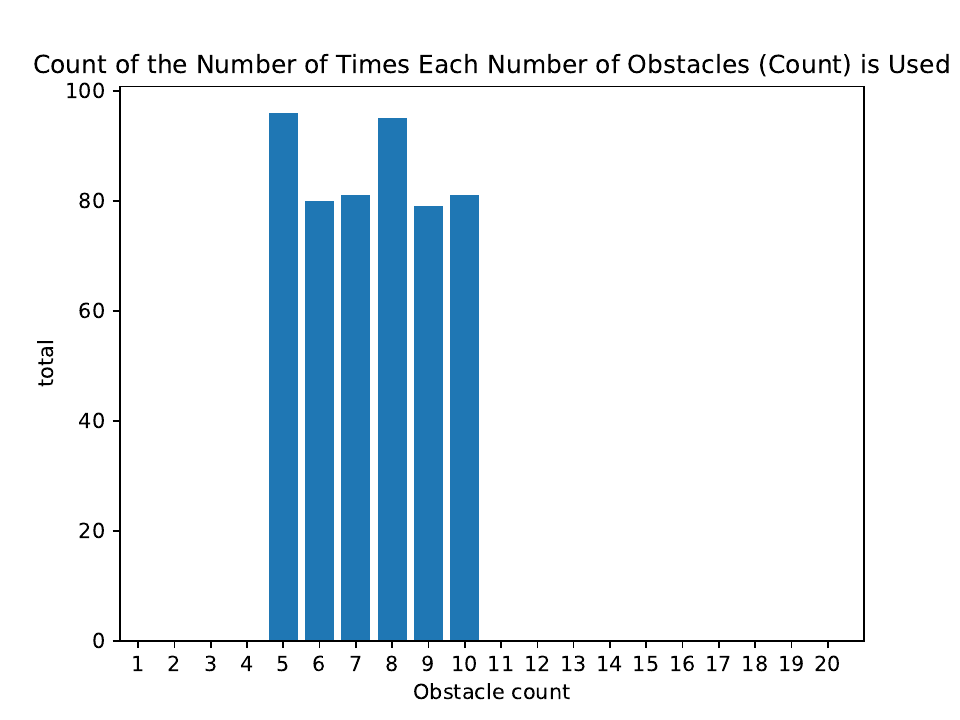}}
     {Random\_5}
&
\subf{\includegraphics[width=45mm]{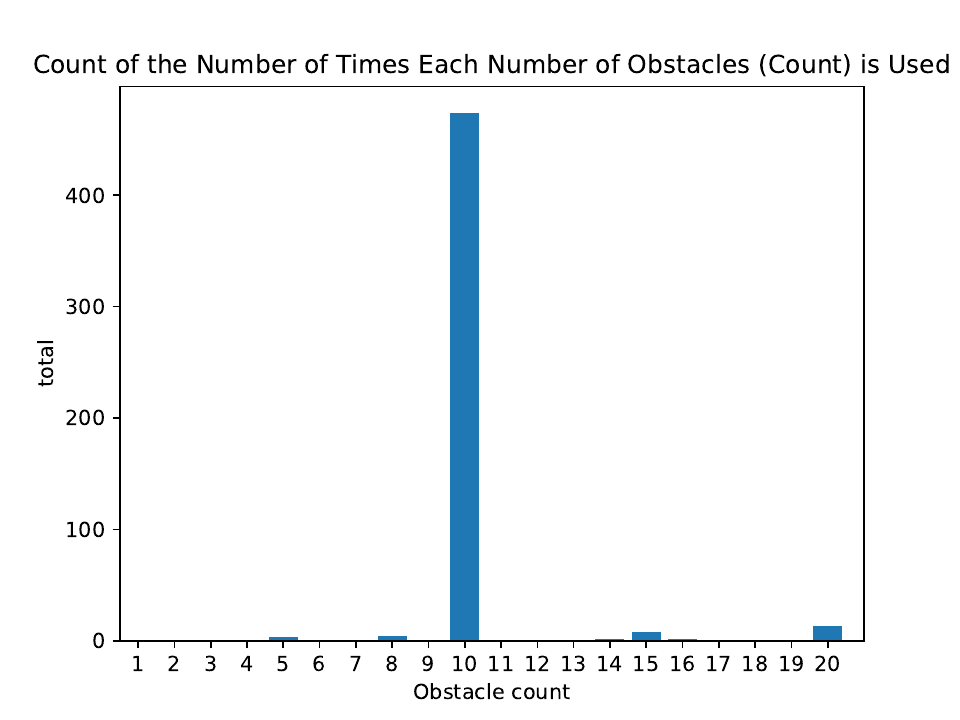}}
     {Agentic}
\\
\end{tabular}
\caption{Chart showing how many times each number of obstacles was selected across all scenarios for each scenario generation method.}\label{fig:numObstacles}
\end{figure}

\begin{figure}[t]
\centering
\begin{tabular}{ccc}
\subf{\includegraphics[width=45mm]{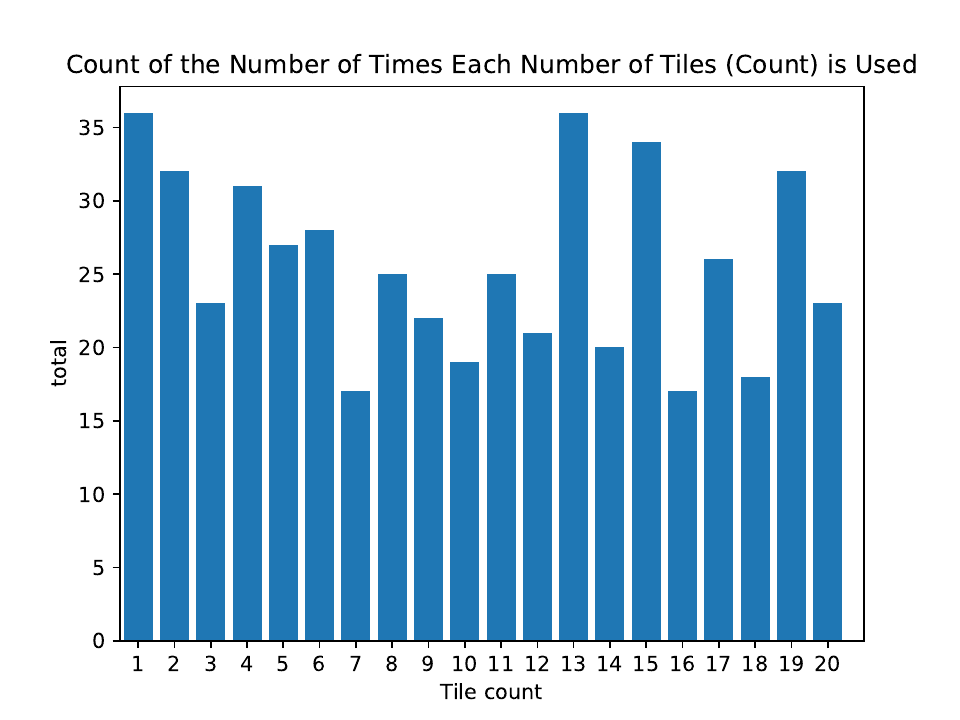}}
     {Random\_1}
&
\subf{\includegraphics[width=45mm]{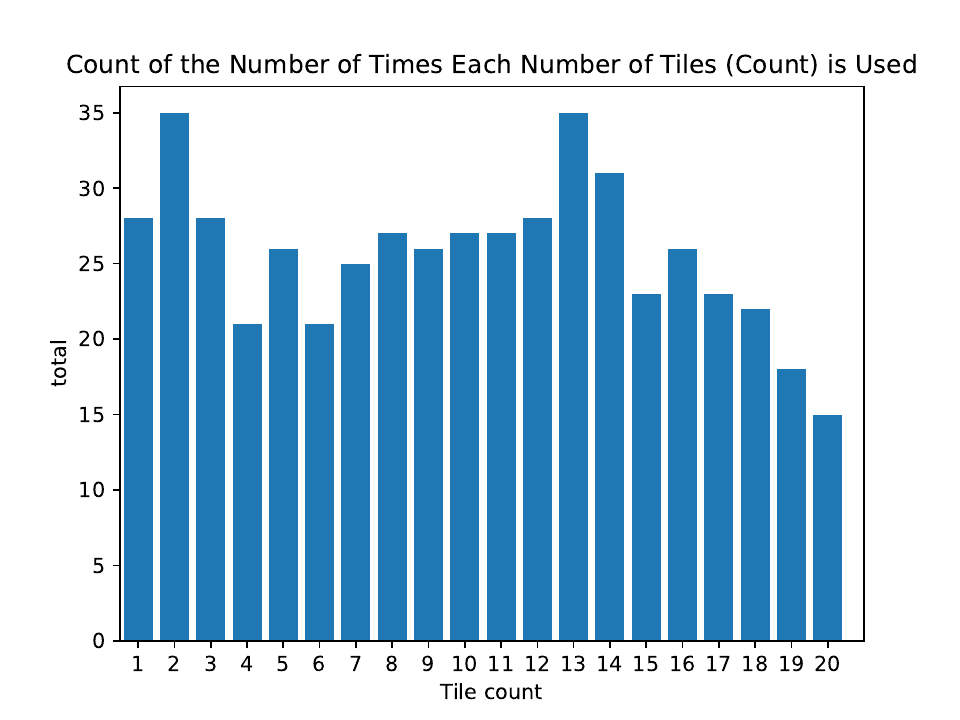}}
     {Random\_5}
&
\subf{\includegraphics[width=45mm]{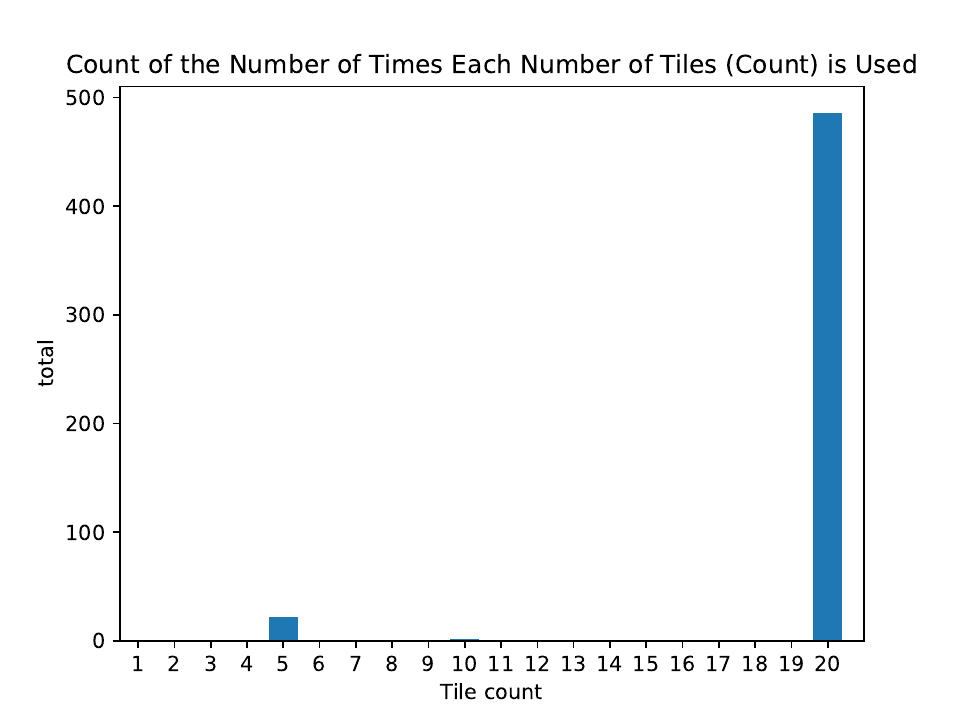}}
     {Agentic}
\\
\end{tabular}
\caption{Chart showing how many times each number (count) of tiles was selected across all scenarios for each scenario generation method.}\label{fig:numTiles}
\end{figure}

\begin{figure}[t]
\centering
\begin{tabular}{ccc}
\subf{\includegraphics[width=45mm]{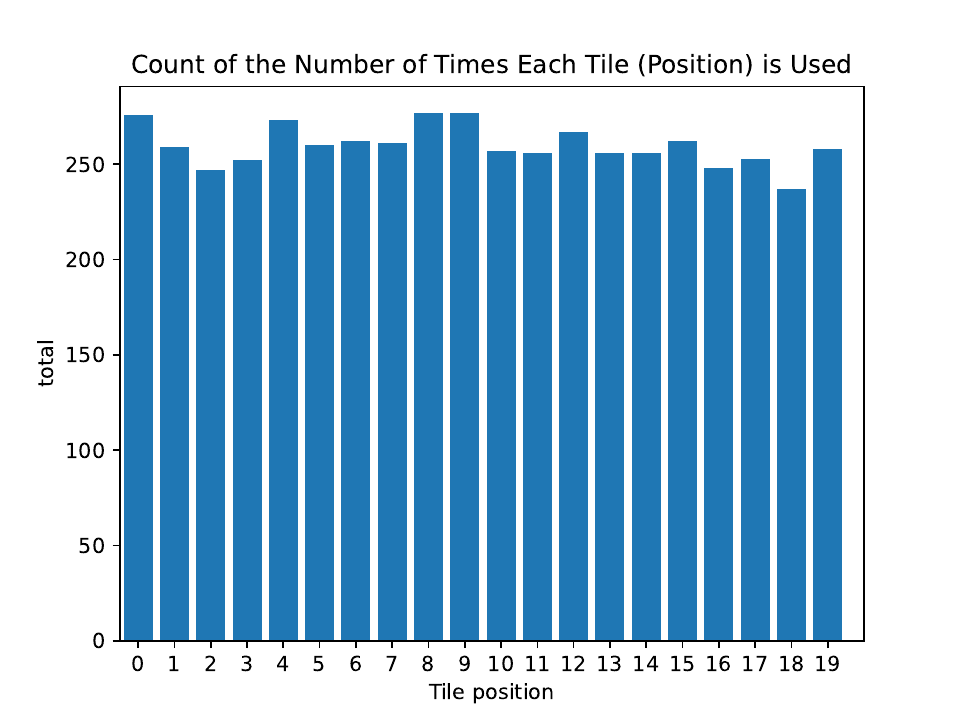}}
     {Random\_1}
&
\subf{\includegraphics[width=45mm]{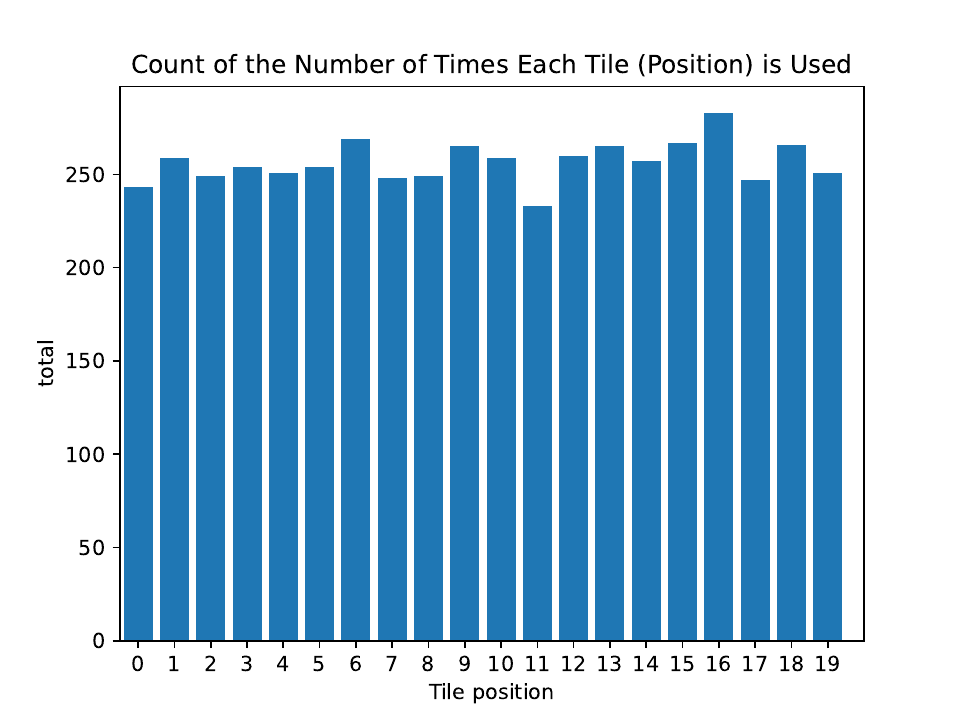}}
     {Random\_5}
&
\subf{\includegraphics[width=45mm]{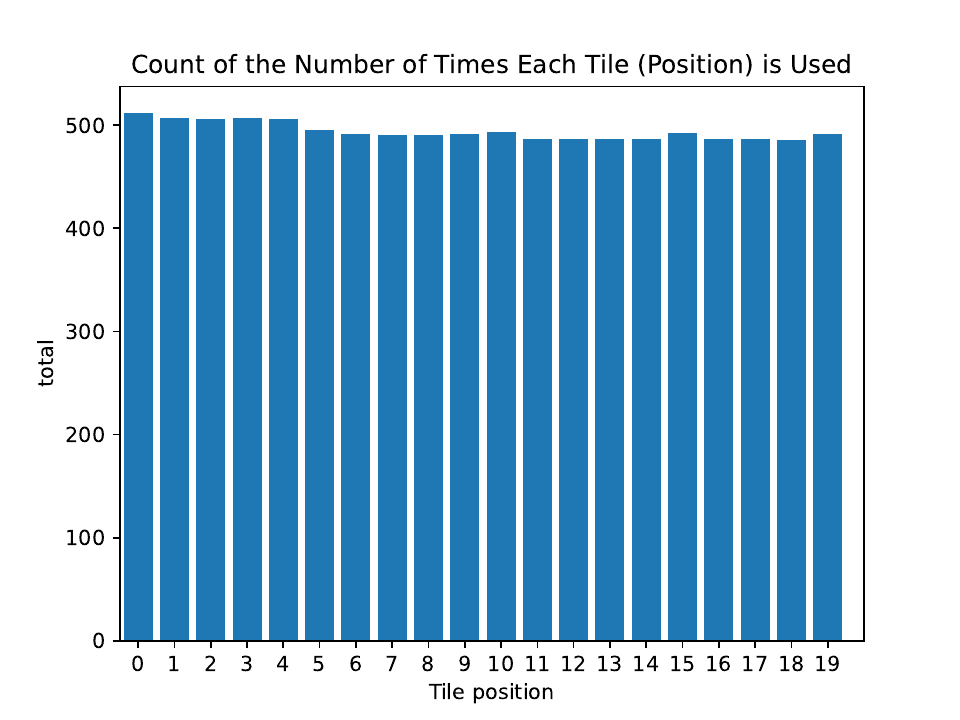}}
     {Agentic}
\\
\end{tabular}
\caption{Chart showing how many times each tile position was selected across all scenarios for each scenario generation method.}\label{fig:tilePositions}
\end{figure}

\begin{figure}[t]
\centering
\begin{tabular}{ccc}
\subf{\includegraphics[width=45mm]{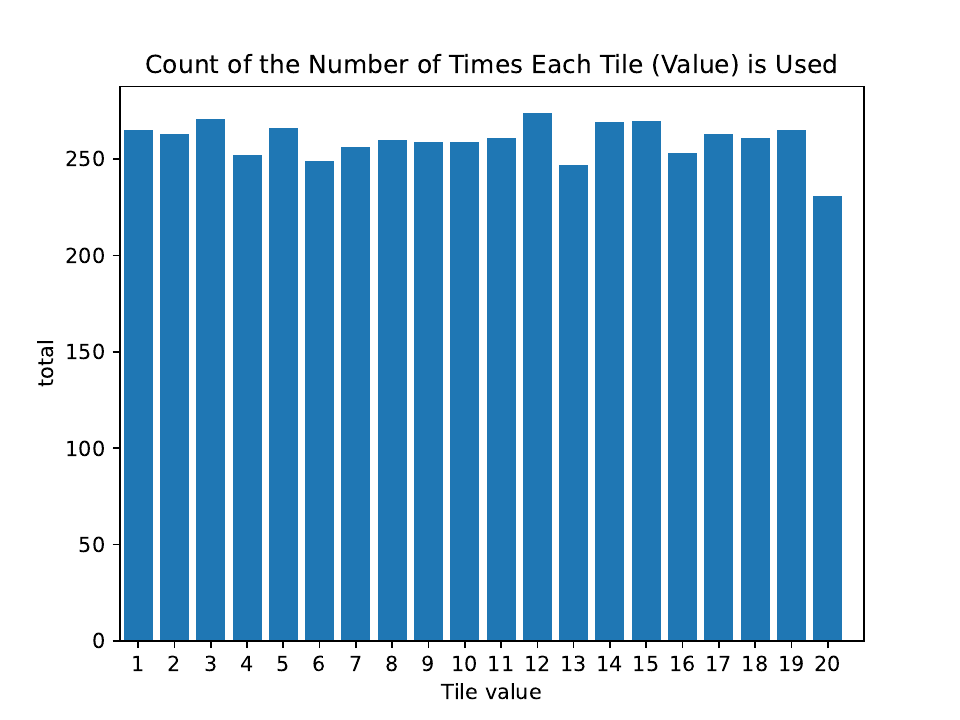}}
     {Random\_1}
&
\subf{\includegraphics[width=45mm]{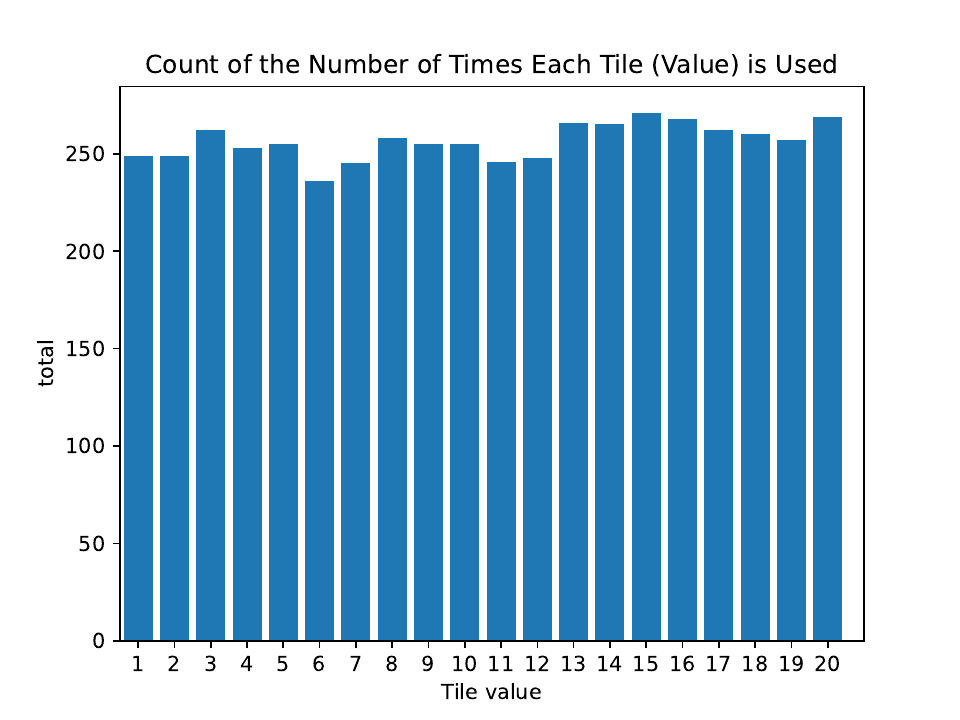}}
     {Random\_5}
&
\subf{\includegraphics[width=45mm]{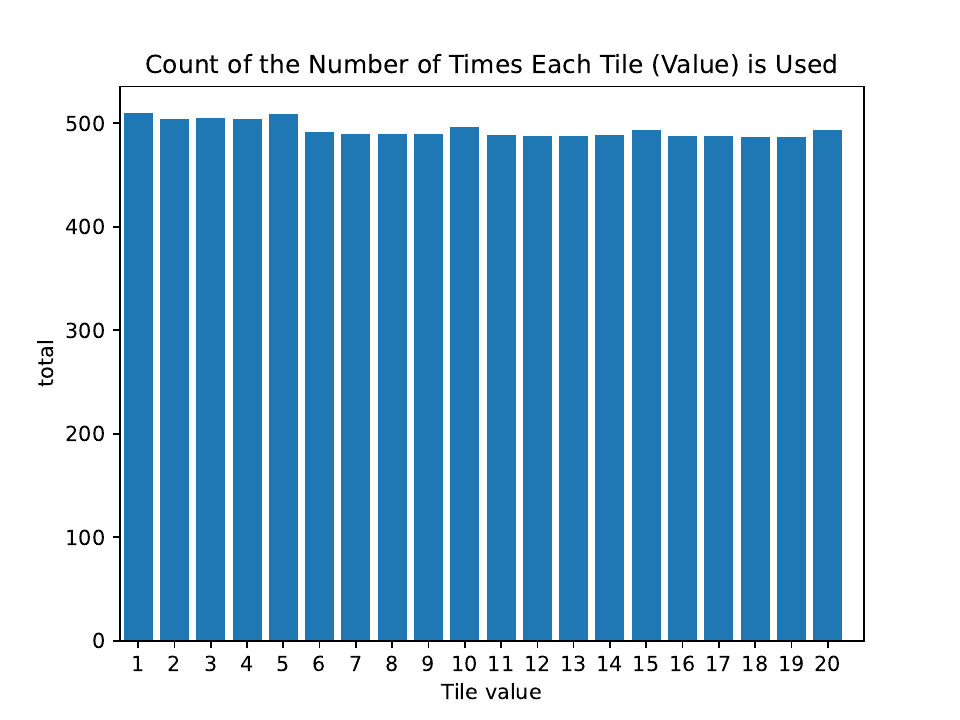}}
     {Agentic}
\\
\end{tabular}
\caption{Chart showing how many times each tile value was selected across all scenarios for each scenario generation method.}\label{fig:tileValues}
\end{figure}

\begin{figure}[t]
\centering
\includegraphics[width=8cm]{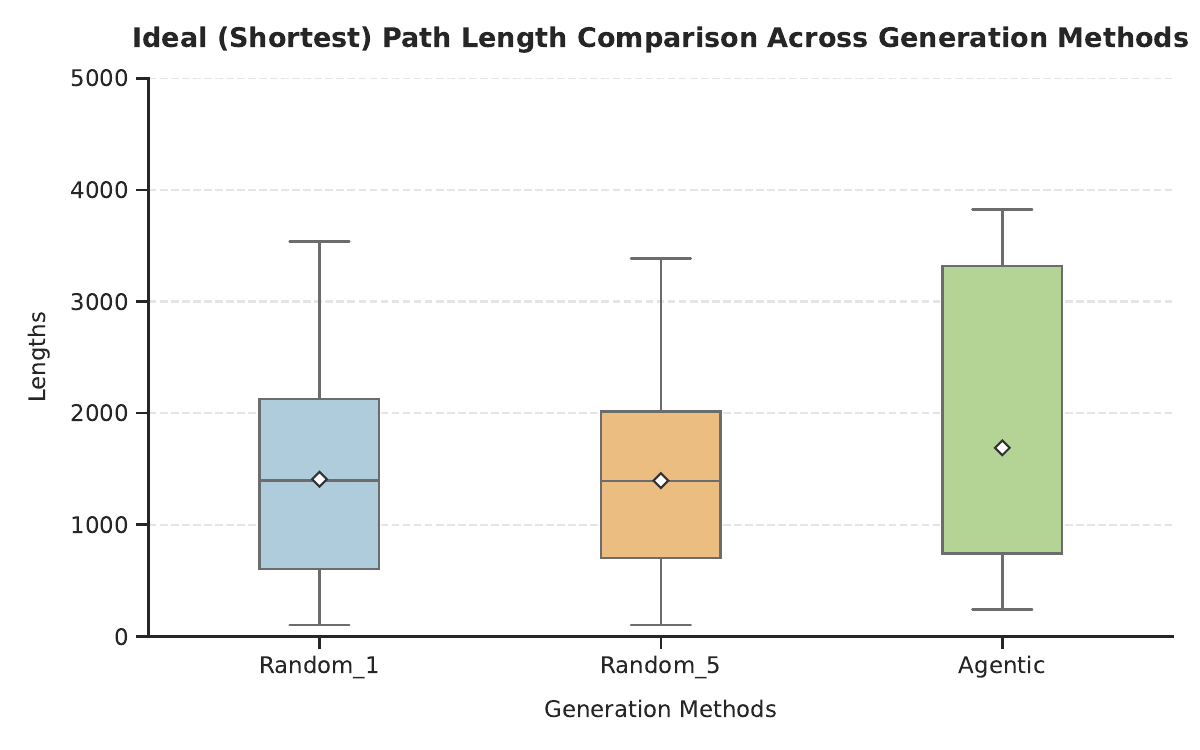}
\caption{Chart showing the ideal path length across all scenarios for each scenario generation method.}\label{fig:idealPath}
\end{figure}

The charts in Figures \ref{fig:heatmap}, \ref{fig:numObstacles}, \ref{fig:numTiles}, \ref{fig:tilePositions} \& \ref{fig:tileValues} demonstrate coverage and completeness. Figure \ref{fig:heatmap} shows the spatial heatmap of obstacle positions (how often each position is selected), Figure \ref{fig:numObstacles} shows the frequency of the number of obstacles selected and Figure \ref{fig:numTiles} shows the number of tiles chosen. Figures\ref{fig:tilePositions} \& \ref{fig:tileValues} illustrate the spread of tile positions and tile values selected by the various scenario generation approaches, respectively. Figure \ref{fig:idealPath} measures the complexity of the scenario (number of steps required to complete) and its spread measures the diversity of scenarios (levels of difficulty).

\subsection{Qualitative Analysis}\label{sec:qualAnalysis}
The following boxes list the name, text description and five evaluation metric scores of the three most critical scenarios that produced the lowest overall reward ($\theta_{JSON_1}, \theta_{JSON_2}, \theta_{JSON_3}$) and the name, text description  and five evaluation metric scores of the scenario $\theta_{JSON_4}$ that produced the highest reward representing a non-critical edge case. We can use these boxes in the safety assurance evidence collection as we can extract information from the text using ``thematic'' analyses, cluster scenarios using these extracted ``themes'' and then examine the various metrics to analyse where the navigation went wrong. For example, ``\textit{dense clusters of obstacles near the agent's starting point}'' in $\theta_{JSON_1}$ and $\theta_{JSON_2}$ or ``\textit{progressively challenging environments}'' both uncover functional gaps with low reward, high path length, multiple collisions and many timeouts (stuck), and visiting the wrong tile which, we have noticed during visual inspection, often coincides with the robotic-agent becoming stuck. ``\textit{Obstacles placed in strategic
positions to block the most direct routes}'' in $\theta_{JSON_4}$ are less challenging as shown by the metrics.
\begin{table}[h!]
\begin{tcolorbox}[width=\textwidth,colback=white,colframe=Lavender!75!black,title={\textbf{$\theta_{JSON_1}$:} Clustered Obstacle Isolation - refined} ,colbacktitle=gray!25,coltitle=black]    
      \textbf{Description:} A challenging scenario with a \textbf{dense cluster of obstacles near the agent's starting point}, requiring the agent to \textbf{find a precise path through the cluster} to reach tiles that are positioned around the room's perimeter. The agent must \textbf{navigate carefully to avoid being blocked}. The tiles are ordered sequentially from 1 to 20 to enforce a natural traversal path and prevent reward hacking. Obstacles near the start have been reduced to ensure a clear initial corridor, improving reliability and reducing crash count.\\
\\
      \textbf{Eval Metrics:}\\
\begin{tabular}{|c|c|c|c|c|}
  \hline
  Reward& Avg path length & Collisions & Timeouts & Wrong Tile\\ \hline
-105.8 & 3802 & 89 & 15 & 7\\
  \hline
\end{tabular}\\
\end{tcolorbox}    
\end{table}
\begin{table}[h!]
\begin{tcolorbox}[width=\textwidth,colback=white,colframe=Lavender!75!black,title={\textbf{$\theta_{JSON_2}$:} Clustered Obstacle Isolation},colbacktitle=gray!25,coltitle=black]    
   \textbf{Description:} A challenging scenario with a \textbf{dense cluster of obstacles near the agent's starting point}, requiring the agent to \textbf{find a precise path through the cluster} to reach tiles that are positioned around the room's perimeter. The agent must \textbf{navigate carefully to avoid being blocked}. The tiles are ordered sequentially from 1 to 20 to enforce a natural traversal path and prevent reward hacking.\\\\
      \textbf{Eval Metrics:}\\
\begin{tabular}{|c|c|c|c|c|}
  \hline
  Reward & Avg path length & Collisions & Timeouts & Wrong Tile\\ \hline
-42.9 & 2674 & 144 & 2 & 5\\
  \hline
\end{tabular}\\
\end{tcolorbox}    
\end{table}
\begin{table}[h!]
\begin{tcolorbox}[width=\textwidth,colback=white,colframe=pink!75!black,title={\textbf{$\theta_{JSON_3}$:} Curriculum Obstacle Navigation}, colbacktitle=gray!25,coltitle=black]  
   \textbf{Description:} A \textbf{progressively challenging environment with reduced obstacle density near the center} to allow navigation between tiles. \textbf{Obstacles are strategically placed to require the agent to learn efficient pathfinding} while maintaining a balance between difficulty and learnability.\\\\
      \textbf{Eval Metrics:}\\
\begin{tabular}{|c|c|c|c|c|}
  \hline
  Reward & Avg path length & Collisions & Timeouts & Wrong Tile\\ \hline
-40.2 & 2536 & 6 & 4 & 10\\
  \hline
\end{tabular}\\
\end{tcolorbox}    
\end{table}
\begin{table}[h!]
\begin{tcolorbox}[width=\textwidth,colback=white,colframe=TealBlue!75!black,title={\textbf{$\theta_{JSON_4}$:} Perimeter Loop with Obstacle Interference}, colbacktitle=gray!25,coltitle=black]  
   \textbf{Description:} Tiles are arranged in a circular path around the perimeter with \textbf{obstacles placed in strategic positions to block the most direct routes}. The agent must navigate around obstacles to visit tiles in order 1-20, requiring complex path planning. Obstacles are randomly scaled and rotated to increase difficulty and prevent shortcutting.\\\\
      \textbf{Eval Metrics:}\\
\begin{tabular}{|c|c|c|c|c|}
  \hline
  Reward & Avg path length & Collisions & Timeouts & Wrong Tile\\ \hline
19.6 & 1118 & 0 & 0 & 0\\
  \hline
\end{tabular}\\
\end{tcolorbox}    
\end{table}
We examine and discuss the quantitative and qualitative results next.

\section{Discussion \& Safety Assurance Case Integration}\label{sec:discussion}
Our local, laptop-based framework aims to offer a flexible, efficient, low-cost method that forms part of the stress-testing process of autonomous systems and generates evidence towards demonstrating the safety compliance of autonomous systems, i.e. that we can argue that $g(AS)$ will be met. We analyse the results of our evaluations from both quantitative and qualitative perspectives to identify the key safety assurance evidence generated. 
\subsection{Quantitative Analysis}
First, we discuss what is revealed by quantitatively analysing the sets of evaluation metrics for the Agentic and random generation approaches.
\subsubsection{Functional Gaps}
The ratio of successful and unsuccessful scenarios (those that timed out or terminated as the wrong tile was visited) are shown in the charts in Figures \ref{fig:success} \& \ref{fig:stackRatios} and demonstrate that Agentic scenario generation finds more critical scenarios than random selection as there are fewer successful runs. The box (inter-quartile range) for the final reward in Figure \ref{fig:reward} is higher (higher lower and upper quartile values) for Agentic showing that runs end with higher reward generally but there are more outliers with very negative rewards than Random\_1 or Random\_5. Random\_5 uncovered one scenario with very negative reward as shown by the yellow circle; random generation can find edge cases but it is by chance. We discuss this further in section \ref{sec:limitations}.  
The regret scores for the three scenario generation are similar as shown in Figure \ref{fig:regret} so the path taken by the robotic-agent compared to the shortest path it could have taken are similar across generation methods. Agentic has slightly higher average and higher lower quartile value but the Random  methods have identified some outliers by chance. Regret does not consider how difficult a scenario is to complete but how optimal the path taken was.  

Figure \ref{fig:tilesVisited} highlights that Agentic causes a higher overall tile visiting failure rate as the upper quartile is lower than 1.0 unlike the Random approaches and the average is lower. The minimum value is higher than Random\_5's minimum though but Agentic has more outliers. It is targetting a spread of edge cases with respect to tile visiting failures.  This demonstrates that using both random generation to quickly generate a large set of scenarios and then the more targeted Agentic generation to supplement these with edge cases and particularly critical edge cases is complementary.

Figure \ref{fig:collisionCDF} shows that Agentic creates scenarios that lead to more collisions and thus uncovers more functional gaps than Random alone. This couples with a much greater spread of average path lengths (Figure \ref{fig:avgPath}) in Agentic scenarios shown by the spread of the inter-quartile range which is larger for Agentic. Agentic has a higher lower quartile, higher average and much higher upper quartile path length than the Random approaches. This suggests that Agentic is finding more complex and diverse spatial layouts that require the robotic-agent to take more steps and are more challenging to complete with more risk exposure. The inter-quartile range of the action distribution entropy (Figure \ref{fig:distribScore}) has lower upper and lower quartiles for Agentic while Agentic has a similar inter-quartile range compared to Random\_1 for action transition entropy in Figure \ref{fig:transitScore}. Random\_5 has a higher lower and upper quartile.  The entropy results for both distributions transition entropy suggest the Agentic layouts are forcing the robotic-agent into selecting actions less equally than Random (from the  distribution entropy charts) but not selecting repetitive actions more than Random\_1 from the transition entropy charts. This means it is uncovering functional gaps in the form of a skewed distribution in the action policy of the RL robotic-agent that Random generated scenarios miss.

\subsubsection{Spatial Analyses}
The charts in Figures \ref{fig:heatmap} show that Agentic generation is targeting a subset of scenarios and does not have the coverage or completeness of Random generation. This is to be expected as we want it to target edge cases specifically.  It tends to place obstacles on the 8 cardinal angles and visual examination of a number of the scenarios running in Unity-3D reveals that Agentic places obstacles in patterns such as circles, t-junctions, blocking lines, blocking arcs and diamonds among others, some of which can be seen in Figure \ref{fig:8-arenas}. From Figures \ref{fig:numObstacles} and \ref{fig:numTiles} it generally picks 10 obstacles and 20 tiles which are the maximum permissible though it does use all values and all locations to place the tiles as shown by Figures \ref{fig:tilePositions} and \ref{fig:tileValues} and often targets patterns of tiles placed in conjunction with the obstacles (as detailed in the text descriptions). More tiles and obstacles will lead to more complex scenarios on the whole. This further supports our assertion that Agentic will provide critical test data and supplement random scenario generation during model training by providing targeted scenarios for uncovering functional gaps. 

Figure \ref{fig:idealPath} also indicates that the complexity of scenarios (minimum path length) is higher. There is more of a spread (higher inter-quartile range) than Random indicating greater scenario diversity.

\subsection{Safety Assurance Case Artefact Synthesis}\label{sec:qualSafetyCase}

The text descriptions of the scenarios produced by the Agentic framework allow a thematic-style analysis which reveals the SOTIF trigger conditions and functional gaps (descriptions of failure modes) and groups scenarios by their descriptive similarities.
For example, $\theta_{JSON_1}$ \& $\theta_{JSON_2}$ can be clustered under ``\textit{dense clusters of obstacles near the agent's starting point}''. Both scenarios are critical edge-cases so we can feedback that ``\textit{dense clusters of obstacles near the agent's starting point}'' reveals a functional gap in the navigation policy of the robotic-agent as shown by the evaluation metric scores output by our Unity-3D simulation with low cumulative reward, long paths, many collisions and multiple timeouts and wrong tile visits. These scenarios reveal a SOTIF trigger condition as listed in Table \ref{tab:placeholder}. We can compare the physical layouts for similarity by extracting both the spatial layout and text descriptions from the JSON. From these pairs, we can find layout patterns that cause the robotic-agent problems, for example, patterns that cause a high number of collision or taking too many steps (stuck) and timing out. Additionally, simulators such as Unity-3D allow us to collect logs of all actions which we can mine to determine the actions that led to the failure. These evaluations provide direct empirical evidence for our GSN claims under PAS 8800 such as solutions (evidence requirements) SN1.2, Sn1.4 to Sn1.7  using the scenarios generated for Sn1.3 as shown in Figure \ref{fig:gsnLower} demonstrating that functional gaps \{FG\} under specific triggering conditions \{TC\} have been found, tested and catalogued.

\begin{table}[h]
    \centering
    \caption{Hazard Discovery Report: formal mapping table to demonstrate how the generated spatial layouts for $\theta_{JSON_1}$ and $\theta_{JSON_2}$ map to triggering conditions and functional gaps exposed by the simulation's evaluation metrics. We assume that timeouts are a safety risk as the robot is stuck and may oscillate leading to collisions.}
\ra{1.2}
{
\small
    \begin{tabular}{p{3.5cm}p{3.5cm}p{4.0cm}p{4.0cm}}
    \toprule
      JSON Parameter / Asset Configuration   & {ISO 21448 Triggering Condition (TC)} & {Exposed Functional Gaps (FG)} & {Resulting Unsafe Behaviour}\\
      \midrule
     Dense clusters of obstacles near the robot's starting point  & Algorithmic search space saturation: Complex spatial topology  & Planning oscillations ($FG_{plan}$): Local planner path deadlock or trajectory freezing. & Timeouts: Robot stuck, path oscillation, or trajectory abandonment\\
     Tile access blockades (clearance $< w_{robot}$) & Geometric Constraint Boundary: Access to tile blocked requiring robot to reverse & Perception Insufficiency ($SW_{per}$): LiDAR point-cloud sparsity (no backward facing LiDAR) & Collisions with wrong tile or arena walls, or obstacle contact.
    \end{tabular}
}  \label{tab:placeholder}
\end{table}

When we identify functional gaps, we have found unknown unsafe scenarios and moved these scenarios from SOTIF area 3 to area 2 \textbf{known} unsafe scenarios and developed empirical evidence for our GSN claims. We can then decide how to mitigate any known unsafe scenarios. As we used our incremental curriculum learning during PPO training, we can train the RL policy further including using new lessons and feeding the unsafe scenario(s) back to the RL policy training. The scenarios can also be used in curriculum learning to test the policy at each lesson prior to moving to the next lesson to ensure that the RL policy  becomes robust to rare but critical  events, moving beyond simple random sampling.


\subsection{Limitations}\label{sec:limitations}
The key limitations we highlight for our agentic AI framework are framed around structural trade-offs and evaluation boundaries.

Smaller LMs are more task-specific so we recommend selecting the correct LM for the job. In this paper, we used a coder LM (qwen3-coder:30b) for generating JSON and a reasoner (nemotron3:33b) for critiquing generated scenarios. As well as task specificity, the smaller LMs have smaller context windows and shorter horizons for tasks. We observed that some generated scenarios contained more than 10 obstacles, as shown in Figure \ref{fig:numObstacles}, despite strict specification in the LM prompts for 1-10 obstacles.  Rather than adding extra steps to the contextual prompt which can degrade the outputs further as per-step reliability for LMs compounds with step count, this problem will be best handled by an error check in the Pydantic validation step (steps 3 and 8 in Figure \ref{fig:part1}). The software stack can then either: loop back to step 1 and generate a new scenario or randomly select 10 from the 20 obstacles. 
Thus, for smaller LMs, we recommend guardrails to catch out-of-distribution errors and then to handle them appropriately, such as in the supporting software stack, without adding extra context to the prompt or extra steps to the LM reasoning.

We propose further investigation to pinpoint why the average cumulative reward inter-quartile range is higher at the end of runs (higher upper and lower quartile compared to Random approaches) as Agentic finds more unsuccessful scenarios so we might expect lower rewards (more penalties). This is our prototype version and we have visually observed a cross-section of the scenarios running in Unity-3D. This has revealed that, for example, we may need to tweak the exploitation process as, occasionally, the exploitation is revising the spatial layout of critical scenarios but making the new scenario easy to reach the goal. This may explain the higher final reward for a subset of the revised scenarios.

With respect to our evaluation boundaries, we acknowledge the sim-to-real gap between testing environments such as simulation and the uncertain and unpredictable open-ended operating environments encountered by autonomous systems. Unity-3D is a high fidelity simulator with more realism than many other simulators \cite{platt2022comparative} but does not create full digital twins (environment facsimile). We have specifically chosen the PPO algorithm to underpin the robotic-agent's navigation as PPO provides the facility to learn in simulation and then transition to learning in the real-world using transfer learning in conjunction with incremental curriculum learning. Our scenario generation framework can create scenarios for real-world testing as the scenarios can be laid out using the JSON specification in step 10 in Figure \ref{fig:part2}. The JSON specification described in section \ref{sec:JSONspec} is agnostic of the testing platform (simulation or real-world).

When generating edge cases for scenario-based testing, we have to consider the coverage of the scenarios produced (see section \ref{sec:formal}). There is no absolute coverage metric for our framework as the aim of the framework is to uncover unsafe \textit{unknown} scenarios. The framework currently outputs a broad range of evaluation metrics for each scenario. If we consider the inter-quartile ranges for metrics such as ideal path length and actual path length, the spread of success scores (over 50 repetitious) and couple this with visual inspection of scenarios running in Unity-3D then we can assess the range of scenarios produced. We can mitigate the likelihood of scenario repetition by including other LMs for generating scenarios. Our framework is LM-agnostic through using AI Suite, standardised contextual prompts and knowledge bases, and schema-specified JSON interfaces. We can easily swap LMs. We can even swap LMs during the scenario generation process so we generate a set of scenarios with one LM, and then switch to other LM(s) to generate further scenarios. This should improve the coverage by exploiting the slight variations in LM training and architecture which should increase the variety of generated outputs.

\section{Conclusion \& Future Work}\label{sec:ConclusionFuture}
Developing and operating autonomous systems in the real world, in particular for safety-critical domains, requires us to verify and validate our claim that the system is acceptably safe across its operational design domain. A significant challenge is not arguing the system's safety in standard operating contexts but rather arguing that the system remains acceptably safe in rare and complex edge cases (the tail of the distribution). For example, identifying whether particular spatial layouts of the environment cause the system's navigation stack to fail, i.e., identifying if there is a functional gap in the navigation algorithm with respect to the specific operating context, and what the triggering condition is for any gap. We can then improve the robustness of the AI model used for navigation or incorporate failsafe run-time mitigations to handle these edge case situations.

Generating scenarios for testing by hand is infeasible due to the time required, and random generation cannot target edge cases. Other approaches such as formal modelling, RL or evolutionary algorithms for targeted scenario generation are computationally expensive. Current LLM and agentic frameworks do not target edge cases, require prior knowledge of functional gaps or require human-in-the-loop prompting and verification of scenarios. Our closed-loop local agentic AI platform coupled with a JSON spatial schema interface and run-time evaluation in simulation provides a flexible, low-cost framework for generating edge case scenarios and the associated safety case evidence ($E_{Sim}$). This feeds directly into goal structuring notation (GSN) to help us meet the requirements of safety assurance regulations such as ISO 21448 (SOTIF) and PAS 8800 for autonomous systems. 

Running our agentic framework with local language models (LMs) avoids transmission latency, cloud costs, service outages and privacy issues. This makes safety assurance, specifically targeted scenario-based testing and functional gap discovery, accessible and secure on standard hardware. The agentic framework uses standardised contextual prompts to create an LM-agnostic loop for edge case generation. It translates natural language specifications in the operational design domain (ODD) of the autonomous system into targeted spatial layouts specified using lightweight text-based JSON schemas. These JSON specifications run in a simulator which outputs JSON evaluation metrics. Here we used the high fidelity simulation of Unity-3D but our framework and JSON spatial layout (simulator input) and evaluation metrics (simulator output) schemas are independent of the simulator used.  The evaluation metrics refine and iteratively improve the scenario generation in our closed-loop approach.  The entire closed-loop, from ODD description and requirement parsing, through JSON-encoded scenario generation,  simulation, evaluation and feedback, runs self-contained on a standard laptop. 

ISO 21448 SOTIF mandates that we systematically search for and identify triggering conditions and demonstrate that our system operates within defined safety boundaries and meets its safety goals. When our framework generates a JSON-encoded scenario that causes the robot navigation to violate safety boundaries, such as becoming stuck or not maintaining safe separation from an obstacle, an ISO 21448 area 3 (unknown/unsafe) scenario is identified and converted to area 2 (known/unsafe). The JSON specification coupled with the evaluation metrics serves as a reproducible artefact. LMs are able to generate rich text descriptions of the scenarios such as those shown in Section \ref{sec:results}. By grouping scenarios using the evaluation metrics which point to functional gaps (such as grouping scenarios with high numbers of collisions or scenarios where the robotic-agent repeatedly became stuck and timed out), we can perform a thematic-type analysis of the text descriptions in the cluster to find common themes. These themes highlight the specific triggering condition, $TC$ and we can relate $TC \to FG$. This helps to generate verifiable and traceable empirical evidence 
that functional gaps under specific triggering conditions have been identified and catalogued and provides human-readable analyses of these gaps for safety engineers to improve training of the RL policy or create run-time failsafe mitigations for the robotic-agent.

The framework described here is our prototype version. We are considering the following extensions.

\subsection{Future Work}
The scenario-based testing with our agentic framework has revealed functional gaps in the RL navigation policy. For example, the robotic-agent does not know if it is terminally stuck (boxed in). It will either try and get out until it times out (max steps reached) or barge an obstacle out of the way. To improve the post-hoc investigations, we will analyse the set of evaluation metrics output during simulation or real-word testing to determine the best metrics to capture the quality of scenarios and for supporting the evidence gathering process for safety assurance. We can also use the generated scenarios in RL curriculum learning to improve the RL model's robustness and then retest the updated RL policy using these edge cases and new edge cases to determine if the functional gaps have been fixed. This will help to show that we have reduced the robotic-agent's risk to be as low as reasonably practicable and that it meets its allocated safety requirements. 

Simulators such as Unity-3D produce replay logs that log actions taken and the environment's state. We did not collect the logs here due to disk space limitations as they are very large. However, in future work, collecting the logs will provide forensic information for investigating failures such as collisions, we can investigate the robotic-agent's states, the environment's states and the actions taken prior to a collision and then use this with the thematic-style analyses (section \ref{sec:discussion}) to provide evidence of what went wrong and why. {
\lstset{
  basicstyle=\ttfamily\footnotesize,
  breaklines=true
}
\begin{lstlisting}[keywordstyle={}]
[Text Description] $\to$ [Themes] 
[Themes] + [Spatial Layout] + [Evals] + [Log Evidence] $\to$ Where did it go wrong and why?
\end{lstlisting}

Additionally, we use the logic in Listing \ref{listing:agentic-evals} to determine whether to explore or exploit. By feeding back more metrics and a detailed log in steps 13/14, we can categorise collisions and uncover gaps such as the robotic-agent is 100\% safe in daylight but 0\% safe in Unity-3D ``Night Mode'' or identify that the robot is spinning in circles and stuck when obstacles form a specific shape.

We will evaluate the framework using a different Unity-3D environment to assess the framework's reproducibility and generalisability. Additionally, the Unity-3D Sorter environment focuses on navigating around static obstacles and visiting static waypoints (tiles). Future development will extend our  framework to environments with dynamic obstacles and moveable waypoints. This will require augmenting the JSON schema to support dynamic trajectory specification. The agentic framework can then generate reactive, moving hazards alongside static layouts. 

We will also investigate transferring the generated JSON layouts to physical hardware-in-the-loop testbeds, whether the JSON schema needs augmenting for this task and whether we can overcome the sim-to-real gap.

\section*{Statements and Declarations}
\subsection*{Funding Declaration}
The work was supported by the Centre for Assuring Autonomy (CfAA), a partnership between Lloyd’s Register Foundation and the University of York (\url{https://www.york.ac.uk/assuring-autonomy/}).
\subsection*{Competing Interests}
The authors have no relevant financial or non-financial interests to disclose.
\subsection*{Data Availability}
All data used in this paper were generated using the agentic framework described in the paper.
\subsection*{CRediT Authorship Contribution Statement}
VH: Conceptualisation, Methodology, Data Curation, Software, Validation, Investigation, Formal Analysis, Visualization, Writing - original draft, Writing - Reviewing and Editing.

CP: Conceptualisation, Writing - Reviewing and Editing.

RA: Methodology, Writing - original draft, Writing - Reviewing and Editing
\subsection*{Generative AI statement}
The author(s) declare that no Generative AI was used in the authoring of this manuscript.

\bibliographystyle{elsarticle-num} 
\bibliography{references}






\clearpage
\appendix
\section{}\label{app1}
The Appendix lists:
\begin{description}
    \item[Listing \ref{listing:Unity-setup-YAML}:] the Unity-3D configuration setup, 
    \item[Listing \ref{PPO-setup-YAML}:] the parameters for the PPO RL algorithm used in this research, 
    \item[Listing \ref{curriculum-yaml}:] the parameters for the incremental curriculum learning used to train the PPO RL algorithm, 
    \item[Listing \ref{json-scenario-spec}:] the scenario JSON schema (as JSON, Typescript and Pydantic specifications) which specifies each scenario that is passed to Unity-3D, 
    \item[Listing \ref{listing:envDescription}:] the environment description, 
    \textbf{Listing \ref{listing:sysDescription}:} system description and \textbf{Listing \ref{listing:sysReqs}:} system requirements that combine to form the domain ontology which is used in the language model (agentic) contextual prompt along with the \textbf{Listing \ref{listing:taskDescription}:} task description. 
    \item[Listing \ref{listing:evalJSON}:] Finally, the Appendix lists the evaluations JSON schema (as a Pydantic specification) which contains the evaluation metrics and their scores output for each run of 50 repetitions for each scenario passed to Unity-3D.
\end{description}

{
\small
\begin{lstlisting}[language=Python, caption={Description of the Unity-3D Simulation Environment Setup - the Arena}, label={listing:Unity-setup-YAML}, keywordstyle={}, commentstyle={} ,stringstyle={},  showspaces=false]
Set-up: The robotic-agent is in a circular room with obstacles on the floor and numbered tiles on the perimeter wall. The number of obstacles is random between 1 and 10. The values of the tiles are random between 1 and 20. The obstacles and tiles present in the room are randomized at each episode. When the robotic-agent visits a tile, it turns green.
Goal: Visit all the tiles in ascending order while avoiding collisions with obstacles.
Agents: The environment contains a single robotic-agent
Robotic-Agent Reward Function: 
    -.0004 Existential penalty.
    -.0002 Reversing penalty.
    +1 For visiting the right tile
    -1 For visiting the wrong tile
    -1 For colliding with an obstacle

BehaviorParameters:
    RayPerceptionSensorComponent: 7 : Uses the information from the set of 7 LiDAR ray casts as observations. The rays are cast into the physics world, and the objects that are hit determine the observation vector that is produced.
    Vector Observations: 4 : 2 floats for Position and 2 floats for orientation
    Variable Length Observations : Between 1 and 20 entities (one for each tile) each with 22 observations, the first 20 are one hot encoding of the value of the tile, the 21st and 22nd represent the position of the tile relative to the robotic-agent and the 23rd is 1 if the tile was visited and 0 otherwise.
    Actions: 3 : discrete branched actions corresponding to forward, backward, sideways movement, as well as rotation.
    Stacking: 2 : Stacking refers to repeating observations from previous steps as part of a larger observation. For example, a robotic-agent that stores observations of four steps.This is a simple way to give a robotic-agent limited "memory" without the complexity of adding a recurrent neural network (RNN).

Float Properties: 2
    num_tiles: The maximum number of tiles to sample.
        Minimum: 1, Maximum: 20
    num_obstacles: The maximum number of obstacles to sample.
        Minimum: 1, Maximum: 10

Benchmark Mean Reward: Depends on the number of tiles.
\end{lstlisting}
}

\pagebreak

\begin{lstlisting}[language=Python, caption={PPO robotic-agent training setup specified in YAML}, label={PPO-setup-YAML}]
behaviors:
  Sorter:
    trainer_type: ppo
    hyperparameters:
      batch_size: 512
      buffer_size: 40960
      learning_rate: 0.0003
      beta: 0.005
      epsilon: 0.2
      lambd: 0.95
      num_epoch: 3
      learning_rate_schedule: constant
    network_settings:
      normalize: False
      hidden_units: 128
      num_layers: 2
      vis_encode_type: simple
    reward_signals:
      extrinsic:
        gamma: 0.99
        strength: 1.0
    keep_checkpoints: 5
    max_steps: 15000000
    time_horizon: 256
    summary_freq: 10000
\end{lstlisting}

\pagebreak

\begin{lstlisting}[language=Python, caption={PPO robotic-agent Curriculum Learning Specification in YAML}, label={curriculum-yaml}] 
environment_parameters:
  num_tiles:
    curriculum:
      - name: Lesson0 # The '-' is important as this is a list
        completion_criteria:
          measure: reward
          behavior: Sorter
          signal_smoothing: true
          min_lesson_length: 100
          threshold: 1.0
        value: 2.0
      - name: Lesson1
        completion_criteria:
          measure: reward
          behavior: Sorter
          signal_smoothing: true
          min_lesson_length: 100
          threshold: 1.5
        value: 4.0


...

      - name: Lesson7
        completion_criteria:
          measure: reward
          behavior: Sorter
          signal_smoothing: true
          min_lesson_length: 100
          threshold: 5.0
        value: 16.0
      - name: Lesson8
        completion_criteria:
          measure: reward
          behavior: Sorter
          signal_smoothing: true
          min_lesson_length: 100
          threshold: 6.5
        value: 18.0
      - name: Lesson9
        value: 20.0
\end{lstlisting}

\pagebreak

\begin{lstlisting}[language=Python, 
caption={JSON Scenario Specification (Schema).}, label={json-scenario-spec},keywordstyle={}, commentstyle={}, stringstyle={}, showspaces=false, showstringspaces=false]
{
  "type": "object",
  "properties": {
    "scenarioName": { "type": "string" },
    "description": { "type": "string" },
    "roomRadius": { "type": "number", "default": 20.0 },
    "episodeMaxSteps": { "type": "integer", "default": 5000 },
    "agentPosX": { "type": "number" },
    "agentPosZ": { "type": "number" },
    "agentRotation": { "type": "number" },
    "tiles": {
      "type": "array",
      "items": {
        "type": "object",
        "properties": {
          "oneHotID": { "type": "integer" },
          "relativePos": { "type": "integer" }
        },
        "required": ["value", "index"]
      }
    },
    "obstacles": {
      "type": "array",
      "items": {
        "type": "object",
        "properties": {
          "x": { "type": "number" },
          "z": { "type": "number" },
          "scaleX": { "type": "number" },
          "scaleZ": { "type": "number" }
        },
        "required": ["posX", "posZ"]
      }
    }
  },
  "required": ["scenarioName", "tiles", "obstacles"]
}
\end{lstlisting}

\begin{lstlisting}[language=Python, 
caption={Typescript specification for a scenario.}, label={ts-spec},keywordstyle={}, commentstyle={}, stringstyle={}, showspaces=false, showstringspaces=false]
ts_interface = 
interface MLAgentsScenario {
  scenarioName: string;
  description: string;
  roomRadius: number; // default 20.0
  episodeMaxSteps: number; // default 5000,
  agentPosX: number;
  agentPosZ: number;
  agentRotation: number;
  tiles: { oneHotID: number, relativePos: number }[];
  obstacles: { x: number, z: number, scaleX: number, 
               scaleZ: number }[];
}
\end{lstlisting}

\pagebreak

\begin{lstlisting}[language=Python, 
caption={Pydantic Scenario Schema for Validating JSON-encoded Scenario Outputs}, label={pydantic-spec}, showstringspaces=false ]
]
class Tile(BaseModel):
    oneHotID: int
    relativePos: int

class Obstacle(BaseModel):
    x: float
    z: float
    scaleX: float = 1.0
    scaleZ: float = 1.0

class MLAgentsScenario(BaseModel):
    scenarioName: str
    description: Optional[str] = "No description provided"
    roomRadius: float = 20.0
    episodeMaxSteps: int = 5000
    agentPosX: float = 0.0
    agentPosZ: float = 0.0
    agentRotation: float = 0.0
    tiles: List[Tile]
    obstacles: List[Obstacle]

class Scenarios(BaseModel):
    scenarios: List[MLAgentsScenario]
\end{lstlisting}

\begin{lstlisting}[language=Python, 
caption={The environment description section of the domain ontology, specified as YAML format.}, label={listing:envDescription},keywordstyle={}, commentstyle={}, stringstyle={}, showspaces=false, showstringspaces=false]
# Global Environment Constraints
Environment:
  Shape: Circular
  Radius: 20
  Coordinate_System: Cartesian (x, z)
  Success_Condition: Visit Tiles 1-20 in order

# Class Definitions (The "Blueprints")
Prototypes:
  Tile:
    Components: [oneHotID, relativePos, Visit-State]
    Interaction: Trigger-on-Collision
  Obstacle:
    Properties: [Static, Box-Collider]
    Scale_Range: [1.0, 3.0]

\end{lstlisting}

\begin{lstlisting}[language=Python, 
caption={The system description section of the domain ontology, specified as YAML format.}, label={listing:sysDescription},keywordstyle={}, commentstyle={}, stringstyle={}, showspaces=false, showstringspaces=false]
# System
Agent: Single agent, discrete movement/rotation.
\end{lstlisting}

\begin{lstlisting}[language=Python, 
caption={The system requirements section of the domain ontology, specified as YAML format.}, label={listing:sysReqs},keywordstyle={}, commentstyle={}, stringstyle={}, showspaces=false, showstringspaces=false]
# Constraints:
  1-20 Tiles ordered randomly by (oneHotID, relativePos).
  All tiles have a unique oneHotID index between 1 and 20.
  All tiles have a unique relativePos index between 0 and 19.
  1-10 static Obstacles placed randomly to make navigating as hard as possible.
  The agent always starts at x=0.0, z=0.0
  Ensure that no obstacle is at x=0.0, z=0.0 or overlaps x=0.0, z=0.0 to avoid the agent's start place.
  Ensure that all obstacles are in different positions and that they do not overlap each other.
\end{lstlisting}

\pagebreak

\begin{lstlisting}[language=Python, 
caption={The task description section of the LM prompt, specified as YAML format.}, label={listing:taskDescription},keywordstyle={}, commentstyle={}, stringstyle={}, showspaces=false, showstringspaces=false]
task_description = f"""
## Task
  Generate 4 diverse testing scenarios in strictly valid JSON format following this interface:\n{ts_interface} by calculating exact float coordinates that satisfy the YAML. Focus on an "Edge Case" that would challenge the agent's ability to generalize or handle spatial complexity.
  For each scenario, consider:
    - Entity Layout: Specific placement of obstacles.
    - Entity Order: Specific order of tiles around perimeter of room using relativePos. Vary the relativePos of the tiles - do not just use numerical order.
    - Numerical Complexity: Which tile oneHotIDs are present? Vary the number and oneHotIDs of tiles present. All tile oneHotIDs are between 1 and 20.
    - The "Challenge": Why is this hard for the RL agent?
"""
\end{lstlisting}

\begin{lstlisting}[language=Python, caption={Pydantic Simulation Evaluation Outputs Schema for Validating JSON-encoded Evaluation Metric Outputs}, label={listing:evalJSON}]
class SummaryStats(BaseModel):
    numEpisode: int
    successRate: str
    successScore: int
    timeoutScore: int
    wrongtileScore: int
    avgEpisodeLength: float
    regret: float
    avgPercentageTilesVisited: float
    cumulativeReward: float
    distributionScore: float
    transitionScore: float
    diagnosticVerdict: str

class SpatialAnalysis(BaseModel):
    overallObstacleCollisions: int
    numObstaclesHit: int

class UnityResults(BaseModel):
    scenarioName: str
    scenarioDescription: str
    summaryStats: SummaryStats
    spatialAnalysis: SpatialAnalysis

\end{lstlisting}

\end{document}